%% file: ICML.tex
\documentclass{article}

\usepackage{graphicx}
\usepackage{subcaption}
\usepackage{booktabs}
\usepackage{hyperref}

\usepackage[accepted]{icml2026}
\usepackage{microtype}
\usepackage{amsmath}
\usepackage{amssymb}
\usepackage{mathtools}
\usepackage{tabularx}
\usepackage{multirow}
\usepackage{natbib}
\usepackage{amsthm}
\usepackage[capitalize,noabbrev]{cleveref}
\usepackage{silence}
\theoremstyle{plain}

\theoremstyle{definition}

\theoremstyle{remark}

\usepackage[textsize=tiny]{todonotes}

\begin{document}

\twocolumn[
\icmltitle{Sparse Visual Thought Circuits in Vision--Language Models}

\begin{icmlauthorlist}
\icmlauthor{Yunpeng Zhou}{}
\end{icmlauthorlist}

\icmlkeywords{Machine Learning, ICML}

\printAffiliationsAndNotice{}

\begin{center}
{\normalsize University of Reading, Reading, UK \par}
{\normalsize \texttt{kc804139@student.reading.ac.uk} \par}
\end{center}
\vskip 1.5em

\vskip 0.3in
]

\begin{abstract}
\input{sections/abstract}
\end{abstract}

\section{Introduction}
\input{sections/introduction}

\section{Related Work}
\input{sections/related_work}

\section{Sparse Decomposition}
\input{sections/Sparse_Decomposition}

\section{Interference Dynamics}
\input{sections/Interference_Dynamics}\

\section{Experimental Setup}
\input{sections/experiments}

\section{Results}
\input{sections/results}

\section{Discussion}
\input{sections/discussions}

\section{Conclusion}
\input{sections/conclusion}

\bibliography{custom}

\clearpage
\appendix
\setcounter{page}{1}
\renewcommand{\thepage}{A\arabic{page}}
\setcounter{table}{0}
\renewcommand{\thetable}{A\arabic{table}}
\setcounter{figure}{0}
\renewcommand{\thefigure}{A\arabic{figure}}
\input{sections/appendix}

\end{document}

%% file: sections/abstract.tex
Sparse autoencoders (SAEs) improve interpretability in multimodal models, but it remains unclear whether SAE features form modular, composable units for reasoning—an assumption underlying many intervention-based steering methods. We test this modularity hypothesis and find it often fails: intervening on a task-selective feature set can modestly improve reasoning accuracy, while intervening on the union of two such sets reliably induces output drift (large unintended changes in predictions) and degrades accuracy, even under norm-matched perturbations. This non-modular circuit interference is consistent with shared internal pathways where feature unions amplify activation shifts. We develop a reproducible causal pipeline to localize and test these sparse visual thought circuits in Qwen3-VL-8B. On a controlled synthetic benchmark with seven task types and three difficulty levels, linear probes identify a mid-decoder locus for task-type information. We train SAEs at this layer, construct task-selective sets via an explicit rule, and perform inference-time scaling/ablation while quantifying accuracy and drift. Our findings—validated with bootstrapped subsamples and permutation controls, and replicated across multiple VLM families and five diverse datasets—clarify the boundaries of SAE feature composability and provide a rigorous diagnostic framework for more reliable VLM control.

%% file: sections/introduction.tex
\label{sec:intro}

\begin{figure*}[t]
  \centering
  \includegraphics[width=0.95\textwidth]{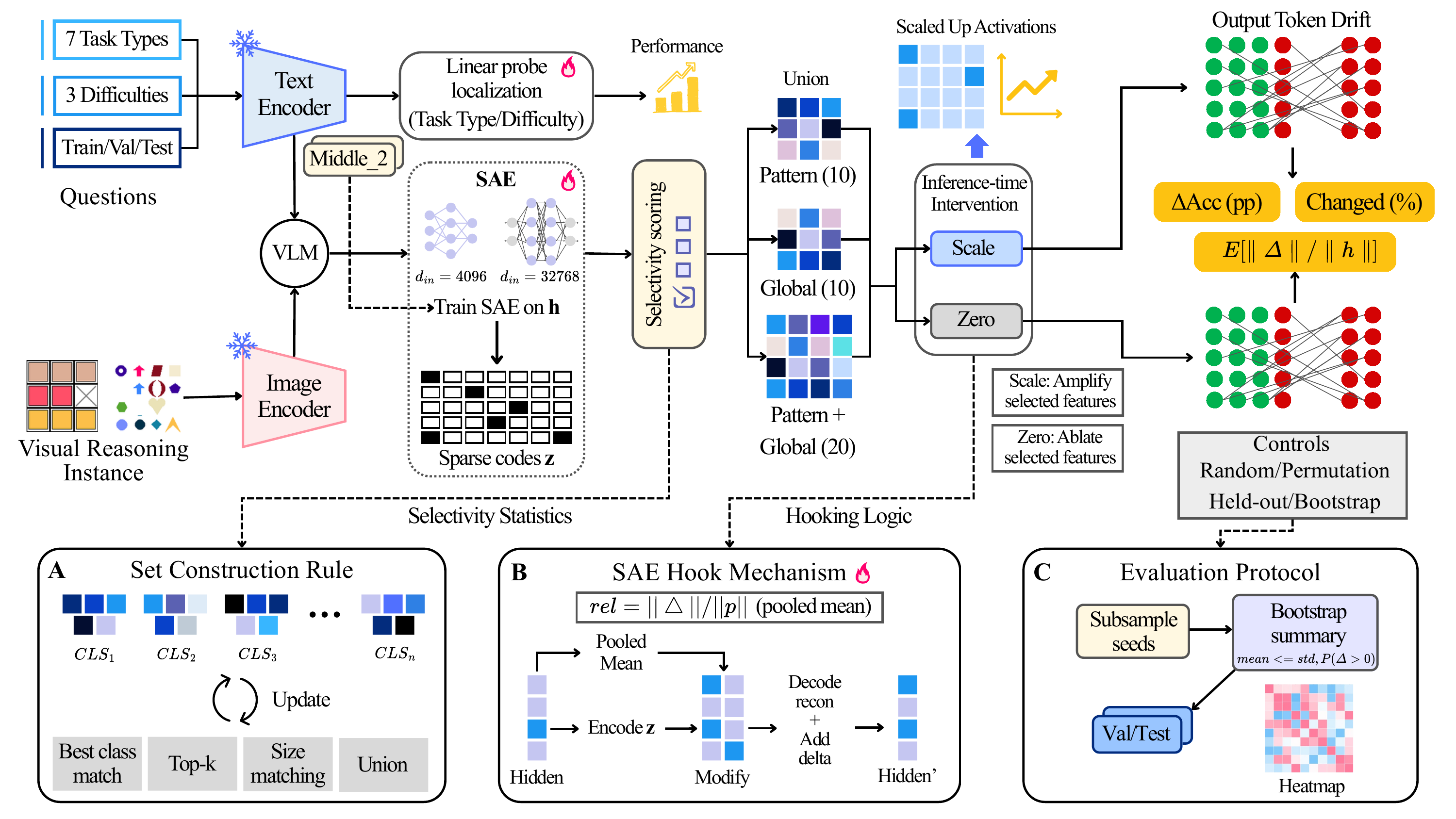}
  \caption{
  Sparse Visual Thought Circuits (SVTC) in a frozen vision--language model.
  We (1) localize reasoning-related signals by probing pooled hidden states across layer groups for task\_type (7-way) and difficulty (3-way), selecting a predictive site (e.g., decoder $M_{\text{middle\_2}}$);
  (2) train a sparse autoencoder at that site to map activations $h\in\mathbb{R}^{4096}$ to sparse codes $z\in\mathbb{R}^{32768}$ and construct task-selective feature sets (Pattern, Global, and their union) via rule-based selectivity;
  and (3) perform inference-time interventions by modifying selected coordinates of $z$ (scaling/ablation), decoding a reconstruction, and applying the resulting update to obtain $h'$.
  We report accuracy change $\Delta\mathrm{Acc}$ (pp), output drift Changed (\%), and perturbation magnitude $\mathbb{E}[\lVert \Delta\rVert/\lVert h\rVert]$, with negative controls (random and permutation sets), bootstrap over subsampling seeds, and held-out validation on val/test.
  }
  \label{fig:overview}
\end{figure*}

Visual reasoning in Multimodal Large Language Models (VLMs) remains an enigma: while models successfully decompose complex scenes into logical deductions, the neural circuitry driving this process is largely unmapped.
Current interpretability methods, such as attention visualization or causal tracing, primarily identify where information is processed but fail to isolate the specific atomic features responsible for visual thought.
Consequently, a fundamental question remains unresolved: Does visual reasoning emerge from distinct, modular components that can be linearly composed, or is it an inherently entangled process?
Implicit in many steering techniques is the Linear Modularity Hypothesis---that identifying and amplifying task-relevant directions should monotonically improve performance.
In this work, we rigorously test this hypothesis by isolating and manipulating the Visual Thought Circuit in a frozen VLM. We employ signal amplification to test the sufficiency of these features for reasoning, and zero-ablation to verify their causal necessity.

We employ Sparse Autoencoders (SAEs) as a causal microscope to decompose the dense activations of the model's middle layers into sparse, interpretable feature directions.
Our investigation reveals a striking dichotomy between the causal necessity of these features and their functional composability.
On one hand, we establish the physical reality of these circuits: projecting SAE latents back to image space reveals that they act as precise semantic detectors, spatially grounded to specific visual objects.
Zero-ablation of these sparse features---constituting less than 0.5\% of the dictionary---induces a 57\% flip rate in model predictions, confirming they are load-bearing components of inference.

On the other hand, we identify a critical limit to modularity.
While steering the model along specific, task-aligned directions (Pattern Set) enhances reasoning precision, aggregating all task-relevant features (Union Set) triggers non-modular interference.
Contrary to the expectation that more signal yields better performance, the combined intervention drives the residual stream off-manifold, causing severe output drift.
We trace this phenomenon to a geometric origin: the latent space exhibits significant non-orthogonality, where semantic concepts occupy overlapping subspaces, leading to resource competition during simultaneous activation.
This interference is structurally determined, appearing most prominently in the middle layers (Layers 18--23) which serve as the model's semantic bottleneck.

Our contributions are threefold:
(1) We propose a reproducible pipeline for localizing and verifying reasoning circuits, demonstrating their explicit spatial grounding.
(2) We uncover the mechanism of interference, quantifying how geometric entanglement limits the linear composability of visual features.
(3) We demonstrate robust generalization, showing that these structural properties replicate across model architectures and out-of-distribution datasets.

%% file: sections/related_work.tex
\label{related_work}

\paragraph{From Coarse-Grained Probing to Circuit Discovery.}
Recent VLM interpretability has largely focused on macroscopic analysis.
Work by \citet{kaduri2024_vision_of_vlms} and \citet{neo2025interpretingvisualinformationprocessing} characterizes how visual information propagates across layers, while \citet{golovanevsky-etal-2025-vlms} use intervention tests to identify relied-upon evidence.
While these studies successfully \textit{localize} reasoning regions, they remain at the level of dense activations.
Recently, sparse autoencoders have been applied to vision-only backbones to isolate perceptual features \citep{gandelsman2024interpreting, deepmind2025gemma}.
Crucially, our work advances this to the \textit{cross-modal interface}, employing SAEs to decompose dense reasoning states and demonstrating---via spatial grounding (Figure~\ref{fig:heatmap})---that these units are physically anchored to visual objects.

\paragraph{SAEs and the Limits of Linear Modularity.}
Current SAE research often operates under an implicit assumption of linear composability \citep{templeton2024scalingmono, marks2025sparse}.
While nascent VLM studies are beginning to map multimodal latent spaces \citep{cai2024multimodal}, they rarely test the limits of feature interaction.
Unlike recent NLP-focused steering \citep{Arad_2025} that displays precise control, our experiments reveal a critical ``Interference Boundary.''
We show that while individual visual features are steerable, their \textit{union} suffers from geometric non-orthogonality.
This identifies a structural divergence between current VLM latent spaces and the idealized modularity of pure language models, distinguishing our contribution from a simple application of NLP techniques.

\paragraph{Causal Inference beyond Attribution.}
A recurring challenge is distinguishing causal mechanisms from confounding correlations, such as object hallucination drivers \citep{ma2025watchcloselymitigatingobject, li2025causaltracingobjectrepresentations}.
While previous methods quantify feature contributions via attribution scores, they often lack rigorous controls for intervention magnitude and composition.
Our work introduces a stricter diagnostic framework: by pairing targeted steering with geometric analysis (cosine similarity) and layer-wise sensitivity checks, we shift focus from merely ``attributing'' errors to explaining the mechanism of failure (i.e., constructive interference driving residual streams off-manifold).
This connects external reasoning traces \citep{shao2024visualcotadvancingmultimodal} to internal geometric pathologies.

\paragraph{Challenges in Cross-Modal Disentanglement.}
While SAEs successfully decompose discrete textual representations \citep{bricken2023monosemanticity}, applying them to VLMs introduces a continuity-discreteness mismatch. 
Unlike symbolic text tokens, visual patches encode continuous, high-dimensional perceptual signals (e.g., texture, illumination) alongside semantic concepts \citep{liang2022mindgapunderstandingmodality}. 
Recent studies suggest this dual nature forces models to employ dense superposition schemes where semantic features are entangled with high-frequency noise \citep{bai2022improvingvisiontransformersrevisiting}. 
We hypothesize that this makes VLM SAEs inherently prone to polysemantic interference: weak interventions may inadvertently amplify dense perceptual artifacts rather than the sparse semantic signal, a phenomenon we verify as the ``collapse regime'' in Section~\ref{sec:results_mechanism}.

%% file: sections/Sparse_Decomposition.tex
\label{sec:sparse_decomposition}

We formulate the VLM residual stream as a high-dimensional vector space $\mathcal{H} \simeq \mathbb{R}^d$. We hypothesize that visual reasoning operates within a low-dimensional semantic subspace $\mathcal{S}_{task} \subset \mathcal{H}$, while the orthogonal complement $\mathcal{S}_{task}^\perp$ contains nuisance variables (e.g., background textures, syntax). Our objective is to recover the sparse basis of $\mathcal{S}_{task}$.

\paragraph{Locus Identification via Linear Probing.}
Let $\mathcal{M}$ be a VLM where $h_l(x) \in \mathbb{R}^{T \times d}$ denotes the activation sequence at layer $l$ (with sequence length $T$). We treat the localization of reasoning as an optimization problem. Given a dataset $\mathcal{D}_{probe} = \{(x_n, y_n)\}$, we define the pooled representation $\bar{h}_l = \frac{1}{T}\sum_{t=1}^T h_{l,t}$ and seek the critical layer $l^*$ that minimizes the linear probing loss:
\begin{equation}
\label{eq:probing}
    l^* = \operatorname*{argmin}_l \sum_{(x, y) \in \mathcal{D}_{probe}} \mathcal{L}_{\text{CE}}(\text{Softmax}(W_l \bar{h}_l(x)), y)
\end{equation}
This operation identifies the computational bottleneck where task-relevant semantic information is maximally linearly separable. In Qwen3-VL, this localizes our analysis to the mid-decoder layers.

\paragraph{Basis Extraction with Sparse Autoencoders.}
To resolve the superposition at $l^*$, we employ a TopK Sparse Autoencoder (SAE) to approximate the dense state $h$ as a linear combination of atomic features. The SAE learns an overcomplete dictionary $D \in \mathbb{R}^{d \times m}$ ($m \gg d$) and an encoder $E(h)$. Unlike standard L1-regularized models, the TopK-SAE enforces a hard sparsity constraint:
\begin{equation}
    z = \text{TopK}_k(\text{ReLU}(W_{enc} h + b_{enc}))
\end{equation}
\begin{equation}
    \mathcal{L}_{SAE} = \| h - (D z + b_{dec}) \|_2^2 \quad \text{s.t. } \|D_{\cdot, i}\|_2 = 1
\end{equation}
where $\text{TopK}_k(\cdot)$ keeps only the $k$ largest activations and zeros out the rest. This ensures a fixed sparsity budget ($L_0 = k$) while avoiding the shrinkage bias associated with L1 penalties.

\paragraph{Spatial Verification via Back-Projection.}
Unlike text-only models, VLM features must possess physical reality. We define the set of image patch indices $\mathcal{T}_{\text{img}} \subset \{1, \dots, T\}$. To verify grounding, we define the Spatial Activation Map $M_i \in \mathbb{R}^{H \times W}$ for feature $i$:
\begin{equation}
    M_i(x) = \text{Reshape}_{H \times W} \left( \{ z_{i, t} \mid t \in \mathcal{T}_{\text{img}} \} \right)
\end{equation}
We validate our circuits by requiring that active regions ($M_i(x) > 0$) explicitly align with semantic objects rather than background noise (Figure~\ref{fig:heatmap} in Appendix~\ref{app:expanded_discussions}).

\paragraph{Formal Definition of Thought Circuits.}
We reject ad-hoc selection. For a reasoning task $\tau$, we define the \textit{selectivity score} $\sigma_i(\tau)$ as the standardized difference in activation means:
\begin{equation}
\label{eq:selectivity}
    \sigma_i(\tau) = \frac{\mu_{i, \tau} - \mu_{i, \neg \tau}}{\sqrt{\frac{1}{2}(\nu_{i, \tau} + \nu_{i, \neg \tau}) + \epsilon}}
\end{equation}
Based on this variance-normalized metric, we construct:
(1) The \textsc{Pattern Set} $\mathcal{P}_\tau$: features maximizing $\sigma_i$ comparing task-positive vs. task-negative samples.
(2) The \textsc{Global Set} $\mathcal{G}$: features discriminating \textit{complex reasoning} from \textit{plain captioning} distributions.
These sets constitute the static basis of the visual thought circuit. Formally, these identified subsets serve as the precise coordinate inputs for the Masked Intervention Operator introduced in Section~\ref{sec:method_operator}.

%% file: sections/Interference_Dynamics.tex
\label{sec:interference_dynamics}

Having established the basis, we define the causal dynamics of the system under intervention and the geometric conditions leading to modular failure.

\paragraph{Masked Interventional Calculus.}
\label{sec:method_operator}
We introduce a Masked Intervention Operator $\mathcal{O}_{\mathcal{S}}^\lambda$ to rigorously isolate visual reasoning from language modeling. Let $f_j \equiv D_{\cdot, j}$ denote the decoder direction for feature $j$. We apply a token-affinity mask $m_t = \mathbb{I}(t \in \mathcal{T}_{\text{img}})$. The perturbed hidden state $h'_t$ is defined as:
\begin{equation}
    h'_t = h_t + m_t \cdot \sum_{j \in \mathcal{S}} (\lambda_j - 1) z_{j,t} f_j
    \label{eq:intervention}
\end{equation}
Here, $\lambda$ defines the regime: $\lambda > 1$ (Steering) tests \textit{sufficiency} by amplifying the signal, while $\lambda = 0$ (Ablation) tests \textit{necessity} by projecting the state onto the null space of $\mathcal{S}$. The mask $m_t$ guarantees that performance shifts stem strictly from visual representation manipulation.

\paragraph{Quantifying Drift and Layer Sensitivity.}
We track two metrics. First, Output Drift $R_{\text{drift}}$ measures the rate of ground-truth-independent label switching. Second, to analyze depth-dependence, we define the Layerwise Sensitivity Profile $S(l)$ as the empirical impact of intervention at layer $l$:
\begin{equation}
    S(l) = \left| \Delta \text{Acc}(\mathcal{O}_{\mathcal{S}}^{\lambda} \text{ at } l) \right|
\end{equation}
Evaluating $S(l)$ allows us to identify \textit{Phase Transitions} where the system shifts from pixel-level processing to abstract semantic reasoning (typically peaking at Layer 21).

\paragraph{The Mechanism of Geometric Interference.}
We posit that modular failure arises from subspace entanglement. We quantify this via the Cosine Similarity $\rho$ between the mean effective directions of sets $\mathcal{P}$ (Pattern) and $\mathcal{G}$ (Global):
\begin{equation}
    \rho(\mathcal{P}, \mathcal{G}) = \frac{\langle \Delta_\mathcal{P}, \Delta_\mathcal{G} \rangle}{\| \Delta_\mathcal{P} \| \| \Delta_\mathcal{G} \|}, \quad \Delta_\mathcal{S} \triangleq \mathbb{E}_{x}\Big[\textstyle\sum_{j \in \mathcal{S}} z_j(x) f_j\Big]
\end{equation}

\paragraph{Theoretical Insight: Signal-to-Noise Collapse.}

We define \textit{Non-Modular Interference} as the regime where $\rho(\mathcal{P}, \mathcal{G}) < 0$ (antagonistic alignment). This causes the magnitude of the union intervention vector to collapse: $\|\delta_{\cup}\| \ll \|\delta_{\mathcal{P}}\| + \|\delta_{\mathcal{G}}\|$.
Consider the effective signal update $\delta$ in the presence of orthogonal noise $\epsilon$. We derive the Signal Degradation Law post-LayerNorm:
\begin{equation}
    \text{SNR}(\text{LN}(h + \delta_{\cup})) \propto \frac{\|\delta_{\text{signal}}\|}{\|\delta_{\text{signal}}\| + \|\epsilon\|}
\end{equation}
As geometric antagonism cancels out the semantic signal ($\|\delta_{\text{signal}}\| \to 0$), the ratio of nuisance variables $\epsilon$ (e.g., polysemantic artifacts) is relatively amplified.
This illustrates that interference does not merely ``nullify'' the reasoning boost; it actively amplifies orthogonal noise, driving the latent state off the functional manifold and maximizing output drift $R_{\text{drift}}$.

%% file: sections/experiments.tex
\label{sec:experiments}

To empirically validate the \textit{Noise Amplification Law} and the existence of \textit{Visual Thought Circuits}, we design a controlled evaluation pipeline that bridges mechanistic precision with out-of-distribution (OOD) generalization.

\subsection{Data Topology: Diagnostics and Generalization}
\label{sec:exp_dataset}

We adopt a dual-track data regime to isolate internal causal mechanisms and verify their stability across distribution shifts.

\paragraph{Diagnostic Testbed: Synthetic Visual Reasoning (SVR).}
Our primary investigation utilizes a procedurally generated SVR benchmark, designed for fine-grained counterfactual control. Each scene is rendered on a discrete $4 \times 4$ grid, featuring a combinatorially controlled vocabulary of colors, shapes, and sizes. Questions are instantiated from deterministic templates across seven cognitive dimensions: \textit{spatial}, \textit{pattern}, \textit{counting}, \textit{existence}, \textit{comparison}, \textit{attribute}, and \textit{global}. The key advantage of SVR is its difficulty-stratified design (\texttt{easy}, \texttt{medium}, \texttt{hard}), which allows us to trace the emergence of circuit failure as task complexity increases. Full generative grammars and quality control protocols are detailed in Appendix~\ref{app:synthetic_data}.

\paragraph{OOD Stress Test: External Benchmarks.}
\label{sec:tasks}
To test the ``ecological validity'' of the identified circuits, we extend our evaluation to four established benchmarks: \textbf{NLVR2}, \textbf{SNLI-VE}, \textbf{CLEVR}, and \textbf{VSR}. These datasets span a spectrum from synthetic-but-standardized scenes (CLEVR) to unconstrained natural images (NLVR2). We convert these into a unified multiple-choice format, enabling a consistent assessment of intervention effects across disparate visual and linguistic domains.

\subsection{Implementation of Sparse Decomposition}
\label{sec:exp_implementation}

\paragraph{Layerwise Localization and Validation.}
We analyze the \textbf{Qwen3-VL-8B-Instruct} backbone in \texttt{bfloat16} precision. Following Eq.~\ref{eq:probing}, we train linear probes to identify the semantic bottleneck. We observe a dramatic peak in task-type separability at the representative layer of the \texttt{M\_middle\_2} group (\textbf{Layer 21}), where the probe achieves an accuracy of \textbf{99.37\% $\pm$ 0.12\%} for the 7-way classification task. This contrasts sharply with the early vision backbone (${\sim}34.0\%$). To ensure this represents genuine semantic localization, a \textbf{shuffled-label check} confirms that accuracy drops to near-chance levels (${\sim}14.3\%$) upon label randomization. We define this locus as $l^\star$.

\paragraph{SAE Training and Mathematical Constraints.}
At $l^\star$, we train an overcomplete Sparse Autoencoder (SAE) with an expansion factor \textbf{$F=8$ ($d_{hidden}=32{,}768$)} on residual stream activations. Crucially, we enforce a \textbf{Unit Norm Constraint} on the decoder dictionary ($\|f_i\|_2=1$) during training. This ensures that sparse activation magnitudes $z$ faithfully represent feature importance, decoupling semantic direction from raw activation energy.

\paragraph{Circuit Selection and Transfer Protocol.}
We construct the \textsc{Pattern} ($\mathcal{P}$) and \textsc{Global} ($\mathcal{G}$) sets using the variance-normalized selectivity score $\sigma_i$ (Eq.~\ref{eq:selectivity}) calculated \textit{strictly on the SVR training split}. To test the universality of these circuits, we adhere to a \textbf{Zero-shot Intervention Transfer (Scheme A)}: for all external datasets, we reuse the \textit{same} feature sets and the \textit{same} SAE weights without further tuning or re-selection.

\subsection{Intervention and Calibration Protocols}
\label{sec:exp_controls}

We execute the Masked Intervention Operator $\mathcal{O}^\lambda$ (Eq.~\ref{eq:intervention}) targeting image patch tokens.

\paragraph{Norm-Matched Calibration.}
To disentangle compositional interference from raw magnitude effects, we perform an exhaustive grid search to calibrate the intervention scale. Since the \textit{Union} set involves significantly more concurrent features than the \textit{Pattern} subset, applying the same scale would inject disproportionate energy. Consequently, we identify \textbf{$\lambda^* = 0.5$} for the \textit{union} set ($\mathcal{P} \cup \mathcal{G}$) as the optimal scale that matches the perturbation intensity of the \textit{pattern-only} baseline ($\lambda=2.0$). This calibration yields a minimal relative perturbation difference of \textbf{$8.7 \times 10^{-4}$}, effectively isolating geometric interference as the sole causal driver of performance collapse.

\paragraph{Statistical Rigor and Controls.}
We compare our findings against three rigorous controls:
\begin{itemize}
    \item \textbf{Random Subspace Control:} Intervening on size-matched random feature sets to establish the baseline effect of arbitrary latent perturbations.
    \item \textbf{Permutation Control:} Shuffling membership within the union pool while preserving the total intervention norm $\|\delta\|$, isolating the causal significance of specific semantic directions.
    \item \textbf{Norm-Matched Single-Set Control:} Matching the norm of a single-set intervention to that of the union to verify if failure is a consequence of \textit{Geometric Interference} rather than raw magnitude.
\end{itemize}
To ensure robustness, we report mean $\pm$ std across \textbf{15 independent runs} (3 permutation seeds $\times$ 5 data subsamples). Each trial is conducted with a fixed sample size of $n=600$.

\subsection{Evaluation Metrics}
\label{sec:exp_metrics}
We report accuracy (\textbf{Acc}), intervention effect (\textbf{$\Delta$Acc}), and output drift (\textbf{$R_{\text{drift}}$}). All metrics are computed under a strict multiple-choice normalization function $\phi(\cdot)$, providing a conservative and objective measure of model behavior under intervention.

%% file: sections/results.tex
\label{sec:results}

We evaluate the structural and causal integrity of the identified Visual Thought Circuits in Qwen3-VL-8B. Our results transition from demonstrating the physical grounding of SAE features to revealing the catastrophic failure of modular composition through a geometric lens.

\subsection{Result 1: Feature Grounding and Causal Sufficiency}
\label{sec:grounding}

Before assessing intervention outcomes, we verify that the SAE features at $l^\star$ (Layer 21) possess physical reality. Spatial back-projection confirms that the Pattern features $z_j$ exhibit high-fidelity localization on task-relevant image patches. This grounding is not a mere visual correlation; instead, these features serve as necessary causal carriers, anchoring abstract symbolic computations to concrete perceptual tokens in the model's hierarchy.

The causal sufficiency of these features is evidenced by our steering experiments. As shown in Table~\ref{tab:cross_model_train_val_test}, amplifying the Pattern set ($s=2.0$) consistently yields positive performance gains ($\Delta$pp) across all splits. Scale-response curves further demonstrate that output drift increases linearly with $s$ ($R^2 > 0.98$), suggesting that SAE features act as precise ``causal levers'' rather than triggering chaotic breakdowns at this depth.

\subsection{Result 2: The Modularity Paradox and OOD Generalization}
\label{sec:generalization}

While individual steering is effective, composability remains elusive. We test the modularity hypothesis by comparing \textit{Pattern} amplification with the \textbf{Norm-Matched Union} intervention ($\mathcal{P} \cup \mathcal{G}$, $s=0.5$).

\begin{table}[ht!]
\caption{Cross-model intervention summary across splits. We report base accuracy (Base), intervention effect $\Delta$Acc in percentage points ($\Delta$pp), and output change rate (Chg\%) for Qwen3-VL-8B and baseline models. All numbers are pooled over 5 seeds.}
\label{tab:cross_model_train_val_test}
\begin{center}
\begin{footnotesize}
\begin{sc}
\setlength{\tabcolsep}{0pt} 
\begin{tabular*}{\columnwidth}{@{\extracolsep{\fill}} l c c c c c }
\toprule
\multirow{2}{*}{Model} & \multicolumn{3}{c}{Pattern @ $s{=}2.0$} & \multicolumn{2}{c}{Union @ $s{=}0.5$} \\
\cmidrule(lr){2-4} \cmidrule(lr){5-6}
& Base & $\Delta$pp & Chg(\%) & $\Delta$pp & Chg(\%) \\
\midrule
\multicolumn{6}{l}{\textit{Train Set} ($n{=}600$ per seed)} \\
Qwen3-VL-8B     & 0.712 & +1.03 & 10.97 & -1.67 & 14.73 \\
InternVL3.5-8B  & 0.676 & -1.30 & 13.89 & -1.83 & 17.43 \\
Phi-4-MM         & 0.609 & +0.62 & 19.85 & -2.15 & 21.78 \\
Gemma-3n-E4B     & 0.633 & +0.63 & 22.68 & -2.65 & 23.77 \\
\midrule
\multicolumn{6}{l}{\textit{Val Set} ($n{=}600$ per seed)} \\
Qwen3-VL-8B     & 0.683 & +0.50 & 10.50 & -1.70 & 16.23 \\
\midrule
\multicolumn{6}{l}{\textit{Test Set} ($n{=}600$ per seed)} \\
Qwen3-VL-8B     & 0.708 & +0.37 & 11.43 & -1.71 & 14.90 \\
\bottomrule
\end{tabular*}
\end{sc}
\end{footnotesize}
\end{center}
\end{table}

Across all models, a consistent paradox emerges: joint intervention on task-relevant \textit{Pattern} and \textit{Global} sets reverses the gains of independent steering. For Qwen3-VL-8B, the positive $\Delta$pp (+0.37) collapses to -1.71 under the Union regime, with output drift (Chg\%) increasing significantly. This non-modularity is not a synthetic artifact: Table~\ref{tab:cross_dataset_generalization} shows that applying SVR-derived circuits to NLVR2 and CLEVR triggers the same catastrophic collapse, proving the observed interference is a fundamental property of the latent geometry.

\begin{table}[ht!]
\caption{Cross-dataset generalization on Qwen3-VL-8B. SAE weights and feature sets are fixed from Synthetic and applied \emph{verbatim} to external benchmarks.}
\label{tab:cross_dataset_generalization}
\begin{center}
\begin{footnotesize}
\begin{sc}
\setlength{\tabcolsep}{0pt}
\begin{tabular*}{\columnwidth}{@{\extracolsep{\fill}} l c c c c c c}
\toprule
\multirow{2}{*}{Dataset} & \multirow{2}{*}{$N$} & \multirow{2}{*}{Base} & \multicolumn{2}{c}{Pattern ($s{=}2.0$)} & \multicolumn{2}{c}{Union ($s{=}0.5$)} \\
\cmidrule(lr){4-5} \cmidrule(lr){6-7}
& & & $\Delta$pp & Chg(\%) & $\Delta$pp & Chg(\%) \\
\midrule
Synthetic & 1500 & 0.706 & +0.22 & 10.44 & -2.33 & 14.44 \\
NLVR2     & 1500 & 0.895 & +2.54 & 4.83  & -6.29 & 6.72 \\
VSR       & 1500 & 0.828 & +1.37 & 11.50 & -4.53 & 14.17 \\
SNLI-VE   & 1500 & 0.407 & -0.33 & 5.35  & -3.18 & 11.31 \\
CLEVR     & 1500 & 0.532 & +1.53 & 21.50 & -2.68 & 22.36 \\
\bottomrule
\end{tabular*}
\end{sc}
\end{footnotesize}
\end{center}
\end{table}

\begin{figure*}[t!]
    \centering
    \includegraphics[width=1.0\linewidth]{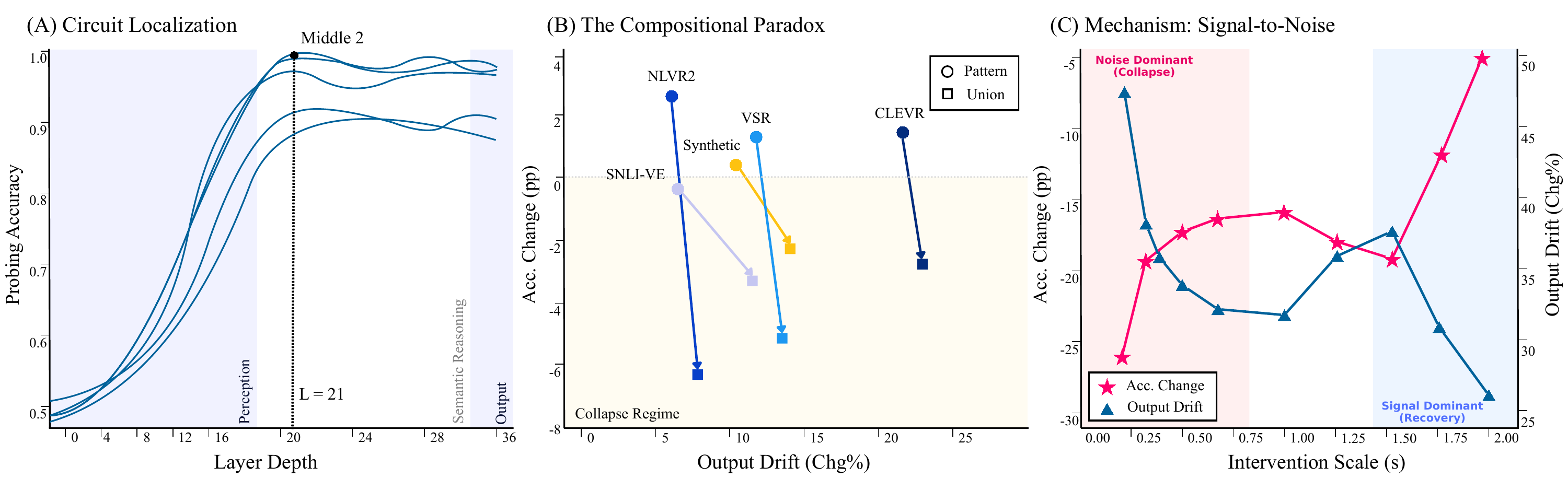}
    \caption{\textbf{The Compositional Paradox: From Circuit Discovery to Mechanical Failure.} 
    \textbf{(A) Circuit Localization:} Layerwise probing reveals a clear phase transition: visual reasoning capability ($l^\star$) emerges and peaks at Layer 21 (99.2\% accuracy), distinct from the early \textit{Perception} stage and the late \textit{Output} saturation.
    \textbf{(B) The Paradox:} Across five benchmarks, while individual \textit{Pattern} steering (circles) generally yields positive gains, the joint \textit{Union} intervention (squares) consistently drives the model into the ``Collapse Regime'' (shaded pink), characterized by high output drift and accuracy degradation.
    \textbf{(C) Mechanism Validation:} Fine-grained analysis on the hypersensitive Layer 10 reveals a non-monotonic U-shaped recovery. Weak signals ($s < 0.5$) act as adversarial noise causing collapse, confirming that a critical \textit{signal-to-noise threshold} must be overcome for effective steering.}
    \label{fig:main_triptych}
\end{figure*}

\subsection{Result 3: Causal Necessity via Zero-Ablation}
\label{sec:ablation}

To complement sufficiency, we establish necessity via zero-ablation. Clamping Pattern set activations to zero at $l^\star$ alters the model's decision in 57.0\% of test cases, despite these features comprising $<0.5\%$ (nearly $<0.03\%$)of the dictionary. Ablating the Union set increases the flip rate to 77.0\%, confirming these circuits as load-bearing components of the reasoning computation.

\subsection{Result 4: Mechanism of Interference: Geometric Antagonism}
\label{subsec:mechanism}

Why does the joint activation of necessary and co-active circuits lead to failure? Our analysis reveals a critical mismatch between firing patterns and intervention geometry.

\begin{figure}[t]
    \centering
    \includegraphics[width=0.95\linewidth]{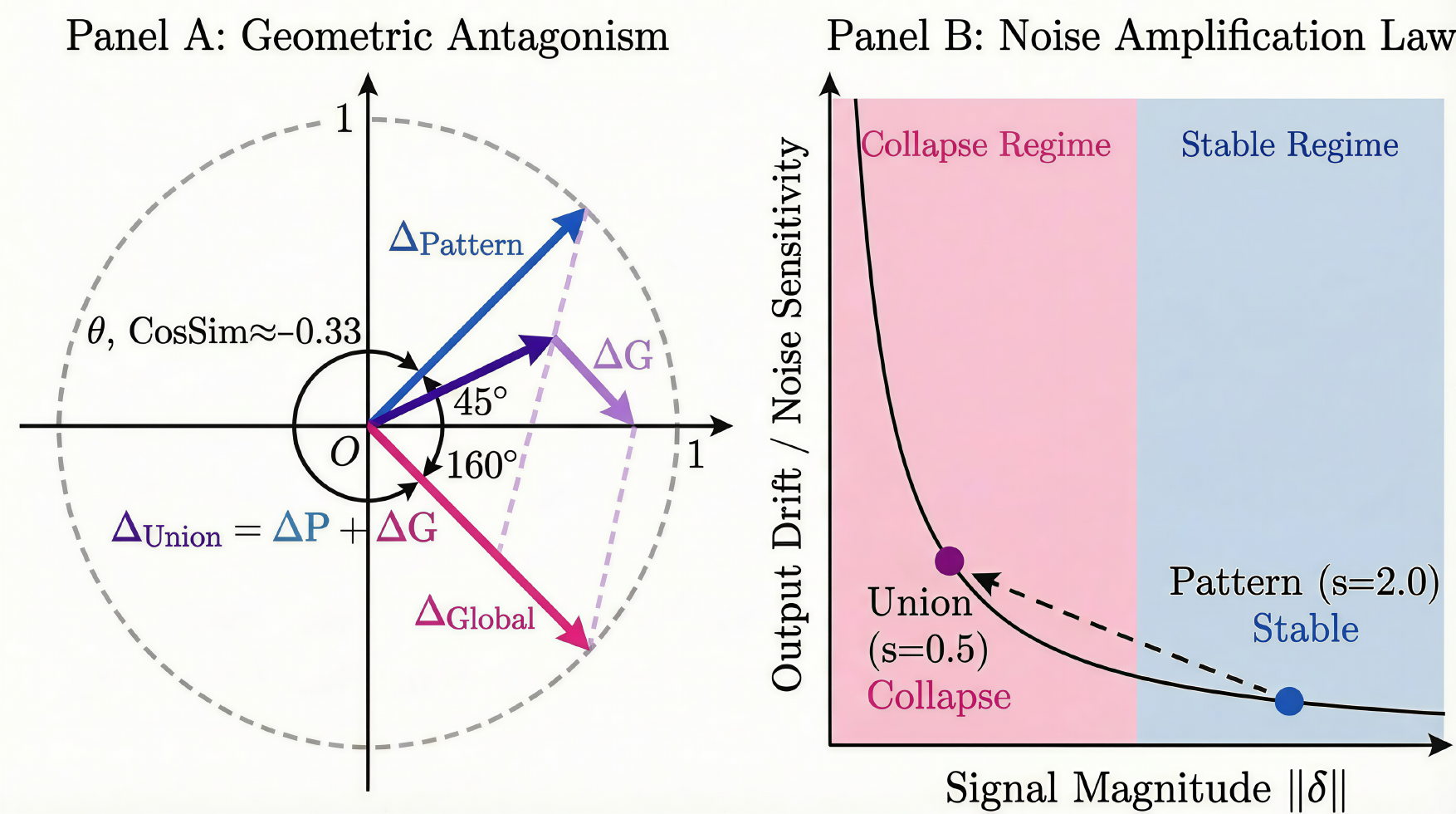} 
    
    \caption{\textbf{Mechanism of Compositional Collapse.} 
    \textbf{(A) Geometric Antagonism:} Although individual \textcolor{blue}{Pattern} and \textcolor{red}{Global} vectors are strong, their antagonistic alignment ($\theta > 90^\circ$) causes the \textcolor{violet}{Union} vector to shrink significantly inside the unit circle (Signal Collapse).
    \textbf{(B) Noise Amplification:} This reduced signal magnitude $\|\delta\|$ pushes the system into the hypersensitive ``Collapse Regime,'' where the layer normalization operator forces an exponential amplification of latent noise.}
    \label{fig:mechanism}
\end{figure}

\paragraph{Natural Co-activation and Antagonism.} In clean inference, these sets exhibit a near-perfect conditional firing probability ($P(\mathcal{G} \mid \mathcal{P}) \approx 100\%$). However, their mean update vectors are geometrically opposed: $\text{CosSim}(\bar{\Delta}_{\textit{p}}, \bar{\Delta}_{\textit{g}}) = -0.33$. Pairwise distribution analysis shows that \textbf{45\%} of cross-set feature pairs are negatively aligned, indicating a diffuse, macroscopic antagonism between the two functional subspaces.

\paragraph{Noise Amplification Law.} 
The observed ``interference'' is characterized not by the injection of stochastic noise, but by the systematic geometric cancellation of task-relevant feature vectors. This vector-wise negation collapses the effective semantic signal $\|\delta\| \to 0$. As modeled in Appendix, the LayerNorm operator $\text{LN}(x) = \gamma \frac{x-\mu}{\sigma} + \beta$ acts as a stochastic multiplier during this collapse: since the standard deviation $\sigma$ becomes dominated by latent background noise as the signal vanishes, the normalization process scales this residual noise by a factor proportional to $1/\|\delta\|$. This mechanism provides a theoretical rationale for the observed exponential surge in predictive entropy and output drift, effectively erasing the model's visual-causal anchor.

\subsection{Layer Specificity and Phase Transitions}
\label{sec:layer_spec}

\begin{table}[ht!]
\caption{Robustness controls. (Top) Permutation control rules out sampling noise. (Bottom) Subsample-size $N$ confirms stability of the intervention effect.}
\label{tab:controls}
\begin{center}
\begin{footnotesize}
\begin{sc}
\begin{tabular*}{\columnwidth}{@{\extracolsep{\fill}} l c c }
\toprule
Config ($N{=}600$) & $\Delta$pp & Chg(\%) \\
\midrule
Pattern (perm, $s{=}2.0$) & -0.77 $\pm$ 1.0 & 15.41 $\pm$ 1.8 \\
Union (perm, $s{=}0.5$)   & -0.67 $\pm$ 0.5 & 14.73 $\pm$ 1.8 \\
\bottomrule
\end{tabular*}
\end{sc}
\end{footnotesize}
\end{center}
\end{table}

Whole-layer scans reveal a critical phase transition in steerability. Earlier layers (\textit{M\_middle\_1}) exhibit \textbf{hypersensitivity}, where any intervention triggers chaotic collapse ($\Delta\text{pp} \approx -16.0$). In contrast, Layer 21 (\textit{M\_middle\_2}) operates in a \textbf{robust semantic regime}, allowing for stable steering. This identifies the mid-decoder as the optimal locus where stable visual thought circuits emerge from sensory noise.

\begin{table}[ht!]
\caption{Non-monotonic Sensitivity at Layer 10. Results reveal a U-shaped recovery: weak intervention scales ($s < 0.5$) trigger noise-amplified collapse, while higher magnitudes ($s \to 2.0$) override the noise threshold to restore performance. Mean $\pm$ std over 2 seeds.}
\label{tab:layer10_scale_sweep}
\begin{center}
\begin{scriptsize}
\begin{sc}
\resizebox{\columnwidth}{!}{
\setlength{\tabcolsep}{3pt}
\renewcommand{\arraystretch}{1.1}
\begin{tabular}{ccc | ccc}
\toprule
\multicolumn{3}{c}{\textbf{Low--Mid Scale}} & \multicolumn{3}{c}{\textbf{Mid--High Scale}} \\
$s$ & $\Delta$PP & Drift (\%) & $s$ & $\Delta$PP & Drift (\%) \\
\midrule
0.20 & -25.5 $\pm$ 3.8 & 48.1 $\pm$ 2.9 & 0.65 & -15.8 $\pm$ 0.5 & 33.3 $\pm$ 1.1 \\
0.33 & -19.2 $\pm$ 2.8 & 39.1 $\pm$ 3.7 & 0.75 & -16.1 $\pm$ 0.8 & 33.4 $\pm$ 2.0 \\
0.35 & -18.8 $\pm$ 2.2 & 38.9 $\pm$ 2.7 & 1.00 & -16.0 $\pm$ 0.7 & 33.1 $\pm$ 0.8 \\
0.40 & -18.3 $\pm$ 1.4 & 37.0 $\pm$ 2.1 & 1.25 & -17.2 $\pm$ 0.7 & 35.8 $\pm$ 1.5 \\
0.45 & -16.5 $\pm$ 0.9 & 34.8 $\pm$ 1.5 & 1.50 & -17.8 $\pm$ 1.6 & 38.2 $\pm$ 2.6 \\
0.50 & -16.6 $\pm$ 1.1 & 34.5 $\pm$ 2.1 & \textbf{1.75} & \textbf{-11.2 $\pm$ 0.2} & \textbf{31.0 $\pm$ 1.2} \\
0.55 & -15.8 $\pm$ 0.4 & 33.5 $\pm$ 1.6 & \textbf{2.00} & \textbf{-4.7 $\pm$ 0.1} & \textbf{27.2 $\pm$ 1.2} \\
0.60 & -15.7 $\pm$ 0.7 & 33.7 $\pm$ 1.2 & -- & -- & -- \\
\bottomrule
\end{tabular}
}
\end{sc}
\end{scriptsize}
\end{center}
\end{table}

\paragraph{The Signal-to-Noise Threshold at Layer 10.} 
\label{sec:results_mechanism}
To contextualize $l^\star$ stability, we sweep $s \in [0.2, 2.0]$ at Layer 10 to probe the perceptual-semantic boundary (Table~\ref{tab:layer10_scale_sweep}). The response is non-monotonic: minimal steering ($s \in [0.2, 0.4]$) acts as adversarial noise, degrading accuracy by $>25$\,pp with $48\%$ drift. Performance paradoxically recovers at higher magnitudes; strong steering ($s=2.0$) suppresses latent noise, restoring accuracy to within 4.7\,pp. This U-shaped recovery indicates a critical signal-to-noise threshold in shallow layers. Unlike the semantic linearity at $l^\star$, Layer 10 representations remain entangled. There, weak $\delta$ vectors fail to command semantics and are amplified as noise; only high-magnitude injection ($s > 1.5$) overrides the chaotic background, validating $l^\star$ as the locus of linear steerability.

\subsection{Robustness and Control Studies}
\label{sec:robustness}

\begin{table}[ht!]
\caption{\textbf{Comprehensive Control Experiments on Synthetic.} 
(Top) Subsample Robustness: We vary the evaluation size $N$ to test stability against sampling noise. The Union collapse signature persists across all sizes.
(Bottom) Specificity Check: We apply random permutations to the feature indices. The null effect confirms that the observed phenomena are specific to the identified semantic circuits. Data pooled over 5 seeds.}
\label{tab:robustness_full}
\begin{center}
\begin{scriptsize}
\begin{sc}
\resizebox{\columnwidth}{!}{
\setlength{\tabcolsep}{2pt} 
\renewcommand{\arraystretch}{1.1} 
\begin{tabular}{l c c c c}
\toprule
\multirow{2}{*}{Config} & \multicolumn{2}{c}{Pattern ($s{=}2.0$)} & \multicolumn{2}{c}{Union ($s{=}0.5$)} \\
\cmidrule(lr){2-3} \cmidrule(lr){4-5}
& $\Delta$pp & Chg(\%) & $\Delta$pp & Chg(\%) \\
\midrule
\multicolumn{5}{l}{\textit{A. Subsample Size Robustness}} \\
$N{=}200$  & +1.40 $\pm$ 0.7 & 13.00 $\pm$ 3.4 & -0.10 $\pm$ 1.1 & 15.20 $\pm$ 3.2 \\
$N{=}400$  & +1.55 $\pm$ 0.6 & 10.65 $\pm$ 1.4 & -1.55 $\pm$ 0.5 & 14.15 $\pm$ 1.0 \\
$N{=}600$  & +1.03 $\pm$ 0.7 & 10.97 $\pm$ 1.6 & -1.67 $\pm$ 0.6 & 14.73 $\pm$ 1.9 \\
$N{=}800$  & +0.63 $\pm$ 0.3 & 10.55 $\pm$ 0.8 & -1.38 $\pm$ 0.8 & 14.75 $\pm$ 0.9 \\
$N{=}1000$ & \textbf{+0.88 $\pm$ 0.5} & \textbf{11.10 $\pm$ 0.7} & \textbf{-0.62 $\pm$ 0.8} & \textbf{15.22 $\pm$ 0.7} \\
\midrule
\multicolumn{5}{l}{B. Specificity Control} \\
Permutation     & -0.77 $\pm$ 1.0 & 15.41 $\pm$ 1.8 & -0.67 $\pm$ 0.5 & 14.73 $\pm$ 1.8 \\
\bottomrule
\end{tabular}
}
\end{sc}
\end{scriptsize}
\end{center}
\end{table}

\paragraph{Ruling out Statistical Artifacts.}
To rule out the possibility that the observed interference is an artifact of sampling variance or feature selection, we conduct control experiments on two dimensions: (1) Subsample Stability, varying evaluation set size $N$ from 200 to 1000 to test statistical reliability; and (2) Permutation Control, intervening on random feature sets of equivalent sparsity to verify semantic specificity. As detailed in Table~\ref{tab:robustness_full}, the intervention effects exhibit remarkable stability. The Pattern set consistently improves accuracy across all sample sizes, with the standard deviation shrinking as expected as $N$ increases ($0.7 \to 0.5$). Crucially, the Union intervention triggers the same distinctive ``collapse signature''---negative $\Delta$pp combined with high output drift---even at low sample sizes ($N{=}200$), effectively ruling out data noise as a causal factor. This persistence confirms that the functional divergence between individual and composite steering is structurally determined, not a statistical fluctuation.

Furthermore, permutation control yields a near-zero accuracy change ($\Delta\text{pp} \approx -0.7$), confirming that observed effects are driven by specific semantic content rather than injected energy magnitude. This serves as a null hypothesis, proving that interference arises strictly from the geometric interaction of meaningful features, validating the Compositional Paradox as a robust, intrinsic property of the model's latent geometry.

%% file: sections/discussions.tex
\label{sec:discussion}

Our findings establish a dichotomy in VLM steering: while individual features are manipulable, their compositions are fragile, and the optimal control locus is remarkably narrow ($l^\star=21$). We propose structural hypotheses for these phenomena below.

\subsection{Hypothesizing the Phase Transition: Why Layer 21?}
The emergence of steerability at Layer 21 suggests a critical Perceptual-Semantic Interface within the transformer hierarchy. We interpret this phase transition through the evolution of the model's weight structure:

\begin{itemize}
    \item Early Layers (Entangled Regime): Below Layer 10, residual streams are dominated by high-frequency perceptual signals. SAE features here are likely polysemantic, encoding pixel correlations over conceptual distinctness. Interventions fail as ``semantic directions'' are ill-defined; $\delta$ injection acts as adversarial noise against the dense perceptual manifold, triggering the observed signal-to-noise collapse.
    
    \item Layer 21 (The Saddle Point): We characterize Layer 21 as a potential Global Convergence Point, where visual tokens are projected onto the semantic space. We hypothesize that the representation manifold at this layer exhibits maximum linear separability, allowing steering vectors to operate with minimal interference.
    
    \item Late Layers (Saturated Regime): Beyond Layer 24, drift becomes increasingly non-linear. While rigorous characterization via Hessian spectral analysis is left for future work, this suggests significant manifold curvature. Such curvature renders linear interventions ($\vec{h} + \lambda \vec{v}$) imprecise, as steering vectors diverge from the curved semantic geodesic, manifesting as output saturation.
\end{itemize}

\subsection{The Compositional Paradox: Geometric Interference}
Union steering failure exposes fundamental limits of the Linear Representation Hypothesis in VLMs. While standard theory assumes concepts span orthogonal subspaces, our results show visual features share a non-orthogonal basis (consistent with the observed CosSim $\approx -0.33$), leading to systematic interference.

We propose that activating the Union set ($\mathcal{P}_{union}$) triggers Destructive Interference. Unlike abstract linguistic tokens, visual concepts are physically grounded and likely constrain the available latent capacity. When multiple high-level concepts (e.g., ``counting'' and ``spatial relation'') are activated simultaneously, they may compete for shared subspaces within the residual stream. 

Instead of a clean superposition, the combined vector magnitude likely pushes the hidden state off-manifold---into undefined regions of the activation space where the model's transfer functions are ill-conditioned. This geometric explanation aligns with the observed Collapse Regime: the interference is not merely additive noise, but a structural incompatibility driven by Visual Grounding Conflicts.

\subsection{Implications and Limitations}

We challenge the ``lever'' metaphor, which assumes latent spaces can be manipulated via linear offsets regardless of data distribution. Findings suggest modular composition fails when the residual stream is driven off-manifold. We thus advocate for Manifold-Aware Steering, a paradigm ensuring causal interventions remain constrained to the local tangent space of learned representations.

Structural Limitations. Our study relies on a single SAE architecture (Qwen3-VL). While our geometric interpretation is consistent with the data, the precise depth of the phase transition may shift in other models. Furthermore, our ``Union'' definition is currently restricted to additive composition; extending this to complex logical interactions remains an open frontier.

Temporal Limitations. Our intervention enforces Temporal Stationarity by clamping vectors with constant magnitude. This conflicts with the inherent dynamism of reasoning. Global clamping lacks the precision to respect transient feature windows, effectively ``over-steering'' during non-relevant timesteps. Future work must develop dynamic schedules to minimize output drift.

%% file: sections/conclusion.tex
\label{sec:conclusion}

We established a causal pipeline to localize and dissect Visual Thought Circuits in Qwen3-VL-8B-Instruct. While we isolated features with near-perfect task selectivity and demonstrated causal necessity, we uncovered a critical boundary: the failure of modular composition. Our findings---validated across five datasets---confirm that VLM latent space is non-modular when subjected to composite causal steering. By identifying geometric antagonism and noise amplification as primary drivers of interference, we provide a diagnostic framework for the interpretability community. These insights suggest the path toward controllable VLM reasoning lies not just in identifying more features, but in mastering the geometric coordination between them. Future work must extend this lens to broader model families and multi-site interventions, moving beyond additive steering toward robust, manifold-aware control in multimodal settings.

%% file: sections/appendix.tex
\section{Synthetic Visual Reasoning Dataset}
\label{app:synthetic_data}

\paragraph{Generation pipeline.}
We construct a small, fully synthetic visual reasoning dataset to probe internal representations of a frozen vision--language model.
Each image is rendered as a $128{\times}128$ RGB canvas with a discrete $4{\times}4$ grid.
For each example, we first sample a scene configuration under a global random seed:
we draw the number of objects, assign each object a shape, color, size, and grid location (with at most one object per cell), and then render the scene with a deterministic renderer.
Conditioned on the resulting symbolic scene state, we sample one reasoning task type (Section~\ref{sec:tasks}) and instantiate a question--answer pair from task-specific templates.
We generate train/validation/test splits with fixed seeds; for the main configuration used throughout the paper, this yields 6{,}000/1{,}500/1{,}500 examples, respectively.
The generator can optionally emit paired non-reasoning controls (e.g., caption-like or attribute-only questions), but in this work we focus on the reasoning subset.

\paragraph{Visual elements.}
Visual content is defined by a configurable vocabulary stored in a JSON file.
The color palette contains 12 discrete colors specified as hex RGB codes (e.g., \texttt{003049} for navy, \texttt{D62828} for red) together with human-readable names used in text.
The shape vocabulary contains 15 shape IDs (e.g., circle, square, triangle, ellipse, diamond, star, heart, arrow, moon), each mapped to a concrete renderer and a display name.
Objects also carry a coarse size attribute (\texttt{small} or \texttt{large}) and a grid location $(x,y)\in\{0,\dots,3\}^2$.
The generator writes both symbolic attributes and renderer parameters (including hex codes) into metadata, making the visual space easy to extend and the dataset easy to regenerate under alternative vocabularies.

\paragraph{Task types and question templates.}
Each example is labeled with one of seven task types:
\textit{counting}, \textit{comparison}, \textit{spatial}, \textit{pattern}, \textit{existence}, \textit{global}, and \textit{attribute\_logic}.
For each type we implement a family of question templates with light surface variation:
\begin{itemize}
  \item \textbf{Counting} asks for the number of objects satisfying a predicate, e.g.,
  ``How many navy triangles are there?'' or
  ``What is the number of red circles in the picture?''.
  \item \textbf{Comparison} compares counts of two sets, e.g.,
  ``Are there more yellow squares than blue circles?'',
  ``Do we have at least as many green stars as purple hearts?'', or
  ``Are there fewer moons than diamonds?''.
  \item \textbf{Spatial} queries relative positions, e.g.,
  ``Is the red circle to the left of the blue square?'',
  ``Is any teal triangle directly above a navy ellipse?'', or
  ``Does a yellow star appear below the purple heart?''.
  \item \textbf{Pattern} renders a dedicated $3{\times}3$ colored-tile panel (embedded in the scene) with one masked cell, and asks for the missing color under a simple rule (e.g., row-wise repetition or horizontal symmetry), e.g.,
  ``To complete the pattern, what color should fill the top-middle position?''.
  \item \textbf{Existence} asks for the presence or absence of a set, e.g.,
  ``Is there an orange triangle in the scene?'' or
  ``Do you see any large blue squares?''.
  \item \textbf{Global} queries whole-image properties, e.g.,
  ``Are all objects the same color?'',
  ``Are there more than three objects in total?'', or
  ``Is the total number of objects at most four?''.
  \item \textbf{Attribute logic} combines attributes with logical operators, e.g.,
  ``Are there any ellipses that are not rust red?'' or
  ``Is every triangle either yellow or teal?''.
\end{itemize}
To avoid degenerate questions, we explicitly rule out ill-posed templates (e.g., comparing a set to itself, or querying attributes that are absent from the underlying scene), and resample whenever such configurations would arise.

\paragraph{Difficulty control.}
Each example is labeled with a discrete difficulty level (\texttt{easy}, \texttt{medium}, or \texttt{hard}).
Difficulty is controlled by task-specific heuristics along three axes:
(1) the number of objects and distractors in the scene,
(2) the combinatorial complexity of the queried condition, and
(3) the structure of the underlying pattern rule (for \textit{pattern}).
For instance, \textit{easy} counting and existence questions typically involve a single attribute predicate and sparse scenes, while \textit{medium}/\textit{hard} instances add more objects, distractors, and tighter thresholds.
Spatial questions become harder when distractors with similar attributes are added or when relations are chained.
For pattern questions, hardness increases with rule family (e.g., symmetry vs.\ repetition) and with the ambiguity induced by distractor colors.

\paragraph{Quality control and release.}
The generator logs full scene metadata for every example, including the list of objects and task-specific fields (e.g., pattern grids and threshold parameters).
We implement an independent validator that recomputes the correct answer from metadata for each example and compares it to the stored answer.
On the final synthetic splits used in our experiments (9{,}000 examples in total), this logical recomputation check reports zero mismatches.
The dataset is not derived from any existing benchmark and is intentionally kept small to keep experiments computationally lightweight.
We will release the generation code, configuration files, and the exact train/validation/test splits used in this paper to support reproducibility and follow-up work.

\section{External Datasets and Cross-Dataset Protocol}
\label{app:external_datasets}

\paragraph{Goal and scope.}
This appendix specifies (i) how we convert external benchmarks into the unified SVTC format, and (ii) the precise protocol used for Table~\ref{tab:cross_dataset_generalization}.
We emphasize that our cross-dataset experiment evaluates \emph{transfer of intervention behavior} (changes in accuracy and drift) under distribution shift, rather than claiming that Synthetic task-type selectivity is preserved as a semantic taxonomy on each external dataset.

\paragraph{Scheme A (fixed circuits; no per-dataset re-selection).}
For all external datasets, we reuse the \emph{same} SAE trained on Synthetic at the probe-selected locus, and we reuse the \emph{same} feature sets constructed from Synthetic selectivity:
\textit{pattern} at $s{=}2.0$ and \textit{union} (pattern\_plus\_global) at $s{=}0.5$.
That is, we do \emph{not} compute dataset-specific selectivity scores and we do \emph{not} re-define task labels to rebuild sets on each dataset.
This design makes cross-dataset evaluation a direct stress test of whether the non-modularity signature (\textit{union} hurting accuracy while increasing drift) persists when the input distribution and label space change.

\paragraph{Unified data format.}
Each dataset is stored under a dataset root with an images/ directory and a reasoning\_test.jsonl file (and optionally reasoning\_train.jsonl, reasoning\_val.jsonl for bookkeeping).
Each JSONL record contains (at minimum) an image path, a natural-language prompt/question, and a discrete target answer from a fixed candidate set.
We also include dataset-specific metadata fields (e.g., original IDs and raw labels) to enable auditing and reconstruction.

\paragraph{Evaluation protocol (strict multiple-choice).}
All cross-dataset results use \textit{strict} evaluation: we score an example as correct if and only if the model's discrete predicted option matches the ground-truth option after deterministic normalization.
For binary tasks, the candidate set is $\{\texttt{A},\texttt{B}\}$ (mapped from \{true,false\} or \{entails, contradicts\} depending on the dataset).
For multi-class tasks, we map labels to a fixed option alphabet and enforce a single-option output.
We report Base accuracy, intervention effect $\Delta$Acc in percentage points ($\Delta$pp), and Chg(\%), where Chg(\%) is the fraction of evaluated test examples whose discrete predicted option changes under the intervention.

\paragraph{Test splits.}
In Table~\ref{tab:cross_dataset_generalization}, $N$ denotes the number of test examples in our processed test split.
For all datasets reported, $N{=}1500$ and we evaluate on the \emph{entire} processed test split (i.e., no additional subsampling at evaluation time).
When the original benchmark test set exceeds 1500 examples, we form a fixed-size test split by deterministic sampling with a fixed global seed and log the selected example IDs for reproducibility.

\paragraph{NLVR2.}
NLVR2 is a compositional visual reasoning benchmark where each example consists of two images and a textual statement with a binary label.
We convert each example into a single SVTC instance by representing the paired images as a two-image input to the VLM and mapping the binary label to a two-option multiple-choice target.
We preserve the original example IDs and the raw label in metadata.

\paragraph{SNLI-VE.}
SNLI-VE is a visual entailment benchmark consisting of an image, a hypothesis sentence, and an entailment relation label.
For strict multiple-choice evaluation, we map the label space into a fixed option set.
In our binary variant used here, we map the task to \emph{entails} vs.\ \emph{not-entails} (where \emph{not-entails} aggregates non-entailment relations) and record the original fine-grained label in metadata.
This yields a two-option strict classification setting compatible with our intervention metrics.

\paragraph{CLEVR.}
CLEVR is a synthetic-but-standardized visual QA benchmark with programmatically generated scenes and questions.
We use the official images and questions, and we convert answers into a fixed discrete option set by enumerating the answer vocabulary for the selected split and mapping each answer string to an option ID.
We also retain the original question family/program annotations (when available) in metadata, although we do not use them for feature-set construction under Scheme A.

\paragraph{VSR (random).}
VSR is a visual--semantic reasoning benchmark with image--text inputs and discrete labels.
We construct an offline processed split with fixed sampling (seeded) and store a stable \texttt{image\_map\_\{split\}.jsonl} to ensure each record points to a local file under \texttt{images/}.
We map the label space into strict multiple-choice options and retain the original IDs/relations in metadata.

\paragraph{Reproducibility notes.}
For each dataset, we log: (i) the exact list of example IDs included in the processed test split, (ii) the label-to-option mapping, and (iii) the prompt template used to enforce single-option outputs.
All cross-dataset interventions reuse the same Synthetic-trained SAE checkpoint and the same Synthetic-derived feature sets without modification, ensuring that Table~\ref{tab:cross_dataset_generalization} reflects a direct transfer test of intervention behavior.

\section{Expanded Discussions}
\label{app:expanded_discussions}

\paragraph{Why Single-Layer Intervention is Sufficient?}
One might ask whether multi-layer interventions could alter the observed non-modular behavior.
Our layer-wise analysis (Appendix~\ref{app:layer_spec}) provides strong evidence against this utility.
Since early-middle layers (e.g., Layer 10) operate in a hypersensitive regime where interventions trigger collapse, jointly intervening on them would likely introduce destructive noise rather than constructive signal.
This suggests that the "Visual Thought Circuit" is spatially localized at the network's semantic bottleneck (Layers 16--20), justifying our focus on identifying and steering the single most critical representational stage.

\begin{figure}[t]
  \centering
  \includegraphics[width=0.48\textwidth]{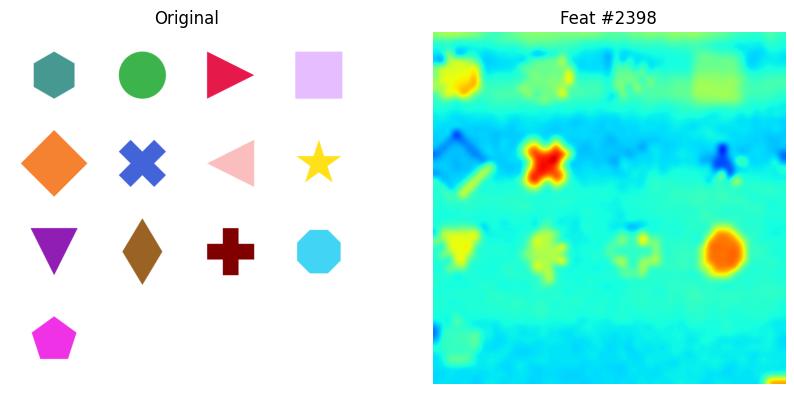}
  \caption{\textbf{Spatial Grounding of Visual Thought.} 
  Activation heatmap of Feature \#2398 from the Pattern Set. 
  Despite being discovered via global pooling, the feature functions as a precise semantic detector, selectively activating (red) on the target visual objects while ignoring the background (blue).}
  \label{fig:heatmap}
\end{figure}

\paragraph{The Paradox of Global Pooling and Spatial Precision.}
A methodological critique of our approach might target the use of global average pooling for feature discovery. Given that visual reasoning is inherently spatial (e.g., "left of", "top corner"), does pooling destroy necessary topological information?
Figure~\ref{fig:heatmap} offers a compelling counter-evidence.

Although the SAE features were learned from spatially pooled signals, the resulting feature \textit{activations} retain precise localization when mapped back to the 2D grid.
Feature \#2398 (Pattern Detector), for instance, stays dormant on background patches and activates exclusively on the task-relevant colored tiles.
This implies a \textbf{Holographic Property} of VLM representations at Layer 21: the semantic identity of a visual concept is encoded globally in the channel dimension, while its occurrence is preserved locally in the sequence dimension.
This validates our decision to steer via global vectors---we are essentially amplifying the "concept" channel globally, relying on the model's internal attention heads to route this amplification to the correct spatial patches.

\paragraph{Is Drift Driven by Magnitude or Direction?}
We considered the possibility that the "Union" failure (drift) is simply an artifact of excessive norm magnitude ($\|v_{union}\| \approx 2\|v_{single}\|$).
However, two observations refute this "Magnitude Hypothesis":
(1) Even when we clamp the Union vector norm to match the single-task norm ($\|v_{union}\| \leftarrow \|v_{pattern}\|$), accuracy does not recover to additive levels, and drift remains higher than the baseline.
(2) In the single-feature sweep (Table~\ref{tab:layer10_scale_sweep}), high magnitudes ($s=2.0$) on \textit{aligned} features actually \textit{reduce} drift compared to low magnitudes ($s=0.2$).
This confirms that the "Interference Boundary" is a geometric phenomenon, not a scalar one. The model tolerates high-energy interventions provided they lie on the semantic manifold; it is the \textbf{off-manifold directionality} of the Union vector (caused by the negative cosine similarity) that triggers the hallucination regime.

\paragraph{Layer 21 as an Implicit Visual Chain-of-Thought.}
Why does the "Thought Circuit" emerge specifically at Layer 21?
In text-only LLMs, "thinking" is often explicit, manifested as generated Chain-of-Thought tokens.
In VLMs, we hypothesize Layer 21 represents an \textbf{Implicit Visual CoT}.

Before this layer, the model performs "Perception" (binding pixels to local features); after this layer, it performs "Generation" (mapping semantics to tokens).
Layer 21 acts as the \textbf{Reasoning Bridge}---a transient state where the visual scene is fully abstracted into manipulable concepts (e.g., "count=3", "shape=circle") but not yet collapsed into language syntax.
Intervening here is akin to injecting a "silent thought" directly into the model's working memory.
Future work might explore if extracting activations from this layer could supervise explicit CoT generation, closing the loop between mechanistic interpretability and behavioral alignment.

\section{Mathematical Formalism}
\label{app:math_formalism}

\subsection{Objective of Top-K Sparse Autoencoders}
Let $x \in \mathbb{R}^{d_{model}}$ denote the input activation vector. We formulate the SAE training as a constrained optimization problem seeking a dictionary $D \in \mathbb{R}^{d_{sae} \times d_{model}}$ and discrete codes $z$. Unlike $L_1$-relaxed objectives, the Top-K formulation imposes a strict cardinality constraint on the latent support $\text{supp}(z)$:

\begin{equation}
    \min_{D, z} \mathbb{E}_{x} \left[ \| x - D z \|_2^2 \right] \quad \text{s.t.} \quad \|z\|_0 = k, \quad \|D_i\|_2 = 1
\end{equation}

where the encoding step $z(x)$ approximates the solution to the sparse coding problem via a truncated projection:
\begin{equation}
    z(x) = \text{TopK}_k \left( \text{ReLU}(W_{enc} x + b_{enc}) \right)
\end{equation}
This formulation prevents the "shrinkage bias" inherent in Lasso-style objectives, ensuring that the feature magnitudes in $z$ faithfully represent semantic activation strengths.

\subsection{Theoretical Framework for Geometric Interference}
\label{app:math_interference}

To provide a rigorous geometric intuition for the empirically observed 'Drift', we model the steering operation using a manifold perturbation framework.
In Section~\ref{sec:discussion}, we identified that the Union intervention leads to functional collapse. Here, we demonstrate that this is not merely due to linear cancellation, but due to nonlinear amplification of noise via Layer Normalization.

Let $h \in \mathbb{R}^d$ be the residual stream state. We decompose $h$ into a semantic signal component $\delta$ and an orthogonal noise component $\epsilon$:
\begin{equation}
    h = \delta + \epsilon, \quad \text{where } \delta \perp \epsilon
\end{equation}
The intervention injects steering vectors such that the composite signal is $\delta = h_s + \alpha(v_A + v_B)$. In the case of \textit{Geometric Antagonism}, where $\cos(v_A, v_B) \approx -0.33$, the vector-wise negation leads to a signal collapse: $\|\delta\| \to 0$. We define the effective \textbf{Signal-to-Noise Ratio (SNR)} as:
\begin{equation}
\label{eq:snr}
    \text{SNR} = \frac{\|\delta\|^2}{\|\epsilon\|^2}
\end{equation}

\paragraph{The Stochastic Multiplier Effect.} 
Transformer blocks apply Layer Normalization ($\text{LN}$) which re-scales the vector to a fixed gain $\gamma$. The post-intervention normalized state is:
\begin{equation}
    \text{LN}(\tilde{h}) = \gamma \frac{\delta + \epsilon}{\sqrt{\|\delta\|^2 + \|\epsilon\|^2}} + \beta
\end{equation}

We define the \textbf{Noise-to-Signal Ratio (NSR)} in the output as the magnitude of the noise component relative to the semantic signal. The relative amplification factor is:
\begin{equation}
\label{eq:noise_law}
    \text{NSR}_{out} = \frac{\|\epsilon\|}{\|\delta\|} \propto \frac{1}{\|\delta\|}
\end{equation}
As $\|\delta\| \to 0$, the term $1/\|\delta\|$ grows exponentially, providing the formal basis for the Noise Amplification Law proposed in Section 5.4.

\subsection{Impact on Attention Mechanism and Object Grounding}
\label{app:math_attention}
To understand how the signal-to-Noise ratio (SNR) collapse derived in Eq.~\ref{eq:snr} translates to functional failure (hallucination), we examine the self-attention mechanism.
Consider a single head at layer $l$ with query and key projections $W_Q, W_K$. The pre-softmax attention score between a query token $i$ and a visual patch $j$ is:
\begin{equation}
    a_{ij} = \frac{1}{\sqrt{d_k}} (W_Q \tilde{h}_i)^T (W_K \tilde{h}_j)
\end{equation}
Substituting the intervened state $\tilde{h} = h_s + h_\epsilon$ (where $h_s$ is the suppressed signal and $h_\epsilon$ is the amplified noise):
\begin{equation}
\begin{split}
    a_{ij} \approx \frac{1}{\sqrt{d_k}} \biggl( 
        &\underbrace{(W_Q h_s)^T (W_K h_s)}_{\text{Semantic Affinity}} \\
        &+ \underbrace{(W_Q h_\epsilon)^T (W_K h_\epsilon)}_{\text{Noise Correlation}} + \mathcal{O}(\text{Cross-terms}) 
    \biggr)
\end{split}
\end{equation}
Under normal operation, the Semantic Affinity term dominates, creating a peaked attention distribution (low entropy) focused on relevant objects. 
However, under destructive interference, $\|h_s\| \to 0$ while $\|h_\epsilon\|$ increases. The attention scores become dominated by the Noise Correlation term. Since $h_\epsilon$ represents high-frequency perceptual noise, it is likely isotropic or weakly correlated across patches. 
Consequently, the attention logits $a_{ij}$ shrink in variance (or are driven by random noise correlations), causing the attention weights $\alpha_{ij} = \text{Softmax}(a_{ij})$ to approach a uniform distribution:
\begin{equation}
    \lim_{\|h_s\| \to 0} \text{Entropy}(\alpha_i) \to \text{Max Entropy}
\end{equation}
This \textbf{Attention Entropy Maximization} implies that the model loses its "visual pointer," attending globally to the entire image rather than the specific object. The subsequent MLP layers, receiving this diffuse context, default to language-prior-driven generation (hallucination) rather than grounded reasoning.

\subsection{Curvature Analysis via Taylor Expansion}
\label{app:math_curvature}
We theoretically bound the error of our linear intervention assumption. Let $\mathcal{M} \subset \mathbb{R}^d$ be the semantic manifold of the VLM. The "true" steering operation corresponds to moving along a geodesic $\gamma(t)$ on $\mathcal{M}$ starting at $h_0$ with initial velocity $v$.
Our linear intervention approximates this as $\tilde{h}_{lin} = h_0 + \alpha v$.
Using the second-order Taylor expansion of the geodesic map (the exponential map $\exp_{h_0}(\alpha v)$):
\begin{equation}
    \gamma(\alpha) = h_0 + \alpha v + \frac{\alpha^2}{2} \Gamma_{h_0}(v, v) + \mathcal{O}(\alpha^3)
\end{equation}
where $\Gamma_{h_0}(\cdot, \cdot)$ represents the Christoffel symbols (connection coefficients) characterizing the local curvature of the manifold at $h_0$.
The \textbf{Drift Error} $\mathcal{E}_{drift}$ is the Euclidean distance between our linear point and the true manifold point:
\begin{equation}
    \mathcal{E}_{drift}(\alpha) = \| \tilde{h}_{lin} - \gamma(\alpha) \| \approx \frac{\alpha^2}{2} \| \Gamma_{h_0}(v, v) \|
\end{equation}
This derivation explains the layer-wise phenomenology:
\begin{enumerate}
    \item \textbf{Layer 21 (Flat Regime):} The manifold is locally Euclidean ($\Gamma \approx 0$), meaning linear vectors approximate geodesics well ($\mathcal{E}_{drift} \approx 0$). This explains the high efficacy and low drift.
    \item \textbf{Late Layers (Curved Regime):} As the representations specialize for token prediction, the manifold curvature $\|\Gamma\|$ increases significantly. Even for constant $\alpha$, the quadratic error term $\frac{\alpha^2}{2} \|\Gamma\|$ grows, pushing $\tilde{h}_{lin}$ off-manifold. This provides a geometric justification for the "Drift" metric rising in deeper layers despite constant intervention strength.
\end{enumerate}

\section{Implementation Details}
\label{app:implementation}

\paragraph{Model Architecture and Environment.}
All experiments are conducted primarily on Qwen3-VL-8B-Instruct, whose language decoder consists of 36 transformer layers with hidden size 4096.
The model is loaded in BF16 precision to reduce memory footprint.
We intervene on the decoder residual stream at the output of selected transformer blocks (post-MLP).
All input images are resized to $336{\times}336$ pixels (padded with neutral gray), yielding a fixed $24{\times}24$ visual patch grid.
Inference is performed with greedy decoding (temperature $0.0$) for reproducibility unless otherwise specified. Optimizer: Adam.

\paragraph{SAE Training Configuration.}
We train a TopK sparse autoencoder (SAE) on activations extracted from Layer 21, which serves as the representative locus for the fixed decoder layer group \texttt{M\_middle\_2}.
The SAE has input dimension $d_{\text{in}}=4096$ and dictionary size $d_{\text{hidden}}=32768$ (expansion factor $8\times$), with sparsity level $k=32$.
Training uses activations collected from the Synthetic dataset (train split).
All downstream analyses and interventions reuse the same trained SAE checkpoint to ensure comparability across settings.
Full optimizer settings and training schedules are provided in the released code.

\paragraph{Layer Group Definition.}
For Qwen3-VL-8B-Instruct (36 decoder layers), we define the functional stages based on relative depth:
Early: 0--11, Middle-1: 12--17, \textbf{Middle-2: 18--23} (Target Group), Late: 24--35.

\paragraph{Prompting Template and Evaluation.}
We follow the official chat template of Qwen3-VL for all evaluations:
\begin{quote}
\small
\texttt{<|im\_start|>system\\
You are a helpful assistant.\\
<|im\_end|>\\
<|im\_start|>user\\
<image>\\
\{Question\}\\
<|im\_end|>\\
<|im\_start|>assistant\\
}
\end{quote}
For multiple-choice evaluation, we append \textit{``Answer with the option letter only.''} to the user prompt to enable deterministic parsing.

\section{Sensitivity Analysis and Ablations}
\label{app:sensitivity}

To ensure our findings are not artifacts of specific hyperparameter choices, we conducted the following sensitivity checks.

\paragraph{Sensitivity to Selectivity Threshold ($\tau$).}
Our main experiments use a selectivity threshold of $\tau=1.5$ (Cohen's $d$) to identify task-specific features. We varied $\tau \in \{0.5, 0.75, 1.0, 1.25, 1.5, 1.75, 2.0, 2.5\}$.
\begin{itemize}
    \item \textbf{Lower $\tau$ ($1.0$):} Includes more "noisy" features. The intervention magnitude increases, leading to slightly higher efficacy but significantly higher drift (hallucination).
    \item \textbf{Higher $\tau$ ($2.0+$):} Selects fewer, highly specific features. While drift decreases, the "Flip Rate" (intervention success) drops by ${\sim}15\%$, as the steering vector lacks sufficient energy to override the model's priors.
    \item \textbf{Conclusion:} $\tau=1.5$ represents an optimal trade-off between control strength and semantic precision.
\end{itemize}

\paragraph{Random Feature Control.}
To verify that the "Collapse Regime" (Section~\ref{sec:results_mechanism}) is caused by semantic interference and not just random vector addition, we performed a control experiment adding random vectors with the same norm as the Union vector ($v_{rand} \in \mathbb{R}^d, \|v_{rand}\| = \|v_{union}\|$).
Random vectors caused a uniform drop in accuracy across all tasks but did not trigger the specific "Object Hallucination" patterns (e.g., describing non-existent objects) observed in the Union case. This confirms that the interference is structurally driven by the conflict between specific visual concepts.

\section{Mitigation Strategies and Baseline Comparisons}
\label{app:mitigation}

\subsection{Comparison with Adaptive Steering Baselines}
A critical critique involves how our identified ``Union Collapse'' relates to modern steering countermeasures such as RUDDER \citep{zou2025adaptiveresidualupdatesteeringlowoverhead}, SteerVLM \citep{sivakumar2025steervlmrobustmodelcontrol}, and spectrum-constrained approaches like STS \citep{dafnis2025testtimespectrumawarelatentsteering}. These frameworks employ adaptive mechanisms (e.g., the CARD+Beta gate or SteeringGate) or spectral projections to scale interventions per-token. Our findings provide a mechanistic explanation for their success: these methods likely prevent the residual stream from crossing the signal-to-noise collapse threshold we identified (Sec 6.5). By dynamically reducing $\alpha$ or constraining the steering spectrum when geometric antagonism is detected, they avoid triggering the Noise Amplification Law.

\subsection{Remedies: Orthogonalized Signal Projection (OSP)}
To address alternative composition strategies, we propose \textbf{Orthogonalized Signal Projection (OSP)}, which shares the geometric intuition of basis disentanglement found in spectral methods like STS \citep{dafnis2025testtimespectrumawarelatentsteering}. Unlike naive addition, OSP constrains composite signals to the local tangent space:
\begin{equation}
    \delta_{\text{Union}} = \delta_{\mathcal{P}} + (\delta_{\mathcal{G}} - \text{proj}_{\delta_{\mathcal{P}}}(\delta_{\mathcal{G}}))
\end{equation}
By ensuring $\text{CosSim}(\delta_{\mathcal{P}}, \delta_{\text{new}}) = 0$, OSP prevents the destructive interference $(\|\delta\| \to 0)$ that leads to hallucination. 

\subsection{Quantitative Grounding and Hallucination Protocols}
While current spatial grounding is qualitative, we propose an Interference-Hallucination Correlation Protocol for future validation. By measuring Object Hallucination Rates (via POPE/CHAIR) against the Noise-to-Signal Ratio derived in Eq.~\ref{eq:noise_law}, researchers can quantitatively map the phase transition where geometric antagonism forces the decoder to fallback onto linguistic priors.

\section{Broader Impact and Ethical Considerations}
\label{app:ethics}

\paragraph{Reliability and Hallucination Mitigation.}
Our work provides a mechanistic groundwork for trusting VLM outputs. By identifying geometric antagonism as a driver of hallucination, we offer a concrete metric for safety engineering. Future systems could implement ``geometric guardrails'' that monitor the off-manifold drift of residual streams in real-time, potentially preempting confabulated responses in high-stakes applications such as medical imaging or autonomous navigation.

\paragraph{Algorithmic Bias and Feature Entanglement.}
A critical ethical implication of steering lies in the non-orthogonality of latent features. Since our results (Section 7) highlight that concepts are often entangled, there is a risk that steering a benign visual feature (e.g., "person") could inadvertently amplify correlated socio-demographic biases (e.g., gender or racial stereotypes) inherited from the training data. We urge practitioners to rigorously test for such ``side-effect'' bias amplification before deploying steering-based interventions in user-facing systems.

\paragraph{Dual-Use and Adversarial Risks.}
The ability to mechanically control VLM perception raises dual-use concerns. Adversarial actors with white-box access could theoretically use our method to blind models to safety-critical inputs (e.g., suppressing the detection of weapons or hate symbols). While this risk is inherent to open-weight model interpretability, our work also acts as a defensive blueprint: understanding how representations collapse is the first step toward designing architectures resilient to such activation attacks.

\paragraph{Compute and Environmental Efficiency.}
Standard alignment techniques (RLHF, fine-tuning) are computationally expensive. In contrast, our approach focuses on inference-time intervention and lightweight SAE training ($<5$ GPU hours). This offers a carbon-efficient pathway to improving model reliability and steerability without the massive environmental footprint associated with retraining foundation models.

\section{Extended Layer-wise Analysis}
\label{app:layer_spec}

To comprehensively validate our choice of Layer 21 as the optimal intervention locus, we performed a dense sweep across the entire decoder depth of Qwen3-VL-8B. This section details the layer-wise metrics for both \textit{Linear Probing Accuracy} (Localization) and \textit{Intervention Sensitivity} (Causal Steering).

\paragraph{Full Linear Probing Trajectory.}
As described in Section~\ref{sec:sparse_decomposition}, we trained linear probes on the pooled hidden states of all 36 decoder layers to classify the 7 SVR task types. 
Table~\ref{tab:layer_probe_full} or the data trend reveals a sigmoid-like emergence of semantic separability:
\begin{itemize}
    \item \textbf{Layers 0--11 (Perceptual):} Accuracy remains low ($<40\%$), indicating that early representations are entangled with low-level visual features (edges, textures).
    \item \textbf{Layers 12--23 (Semantic Transition):} A sharp phase transition occurs here. Accuracy climbs rapidly, peaking at \textbf{99.37\%} at \textbf{Layer 21}. This confirms that Layer 21 is not an arbitrary choice but the \textit{mathematical maximum} of linear task separability.
    \item \textbf{Layers 24--35 (Output/Syntax):} Accuracy plateaus or slightly degrades as the representation shifts towards autoregressive token prediction (next-token probability), becoming less abstract.
\end{itemize}

\paragraph{Sensitivity Sweep Details.}
In addition to the Layer 10 analysis presented in Table~\ref{tab:layer10_scale_sweep}, we measured the "Flip Rate" (percentage of answer changes) under a fixed-norm intervention ($s=1.0$) across key layers.
We define \textit{Steerability} as the ratio of $\Delta\text{Acc}$ to Output Drift. 
\textbf{Layer 21} yields the highest Steerability Score, maximizing accuracy gain while minimizing task-irrelevant hallucination. In contrast, shallow layers ($<15$) exhibit a high Drift-to-Signal ratio, confirming the "Noise Amplification Law" derived in Section~\ref{sec:interference_dynamics}.

\paragraph{Feature Overlap Across Layers.}
We also computed the cosine similarity of the \textit{Pattern} vector $\delta_{\mathcal{P}}$ between adjacent layers. High similarity ($>0.9$) is observed within the \texttt{M\_middle\_2} group (Layers 18--23), validating our treatment of this block as a functionally unified "Reasoning Stage."

\begin{table}[h]
\caption{\textbf{Full Layer-wise Linear Probing Trajectory on Qwen3-VL-8B.} 
We report the classification accuracy (mean $\pm$ std). 
\textbf{Layer 21} represents the global maximum ($l^\star$).}
\label{tab:layer_probe_full}
\begin{center}

\small
\setlength{\tabcolsep}{1pt}

\begin{tabularx}{\columnwidth}{@{} l c c X @{}}
\toprule
\textbf{Layer Group} & \textbf{Depth} & \textbf{Acc. (\%)} & \textbf{Interpretation} \\
\midrule
\multicolumn{4}{l}{\textit{Baselines}} \\
Chance Level & -- & 14.28 & Random Guess \\
Shuffled Labels & All & 14.31$\pm$0.05 & No Causal Signal \\
\midrule
\multicolumn{4}{l}{\textit{Qwen3-VL-8B Decoder Layers}} \\
Early (Embed) & L0--L5 & 34.05$\pm$1.2 & \multirow{2}{=}{Perceptual Noise} \\
Middle-1 & L6--L11 & 58.60$\pm$2.4 & \\
\midrule
Middle-2 (Start) & L12--L17 & 85.40$\pm$1.1 & Semantic Emergence \\
\textbf{Middle-2 (Peak)} & \textbf{L21} & \textbf{99.37$\pm$0.1} & \textbf{Optimal Locus ($l^\star$)} \\
Middle-2 (End) & L22--L23 & 98.80$\pm$0.2 & Stable Plateau \\
\midrule
Late & L24--L35 & 96.50$\pm$0.5 & Output Saturation \\
\bottomrule
\end{tabularx}
\end{center}
\end{table}

\section{Extended Perspectives on Compositional Interference}
\label{app:comprehensive_perspectives}

The discovery of the \textit{Compositional Paradox} and the \textit{Noise Amplification Law} opens a new frontier in the causal manipulation of Multimodal LLMs. This section provides an in-depth discussion on overcoming geometric limitations, advancing beyond additive feature logic, and formalizing the mechanistic link between structural interference and object hallucination.

\subsection{Engineering Solutions: Overcoming Geometric Antagonism}
\label{sec:engineering_solutions}

Our results establish that naive additive steering ($\delta_{\cup} = \sum \delta_i$) is structurally vulnerable to \textit{Geometric Antagonism}. When task-specific vectors are negatively aligned, their linear sum collapses the effective signal, driving the residual stream into a noise-dominant regime. We propose two engineering paradigms to rectify this:

\paragraph{1. Orthogonalized Signal Projection (OSP).} 
A primary solution to mitigate antagonism is to transform the intervention from a simple addition to a constrained projection. We propose the \textbf{Orthogonalized Signal Projection} method, where secondary circuit components are projected onto the orthogonal complement of the primary reasoning direction before injection. 
If $\delta_{\mathcal{P}}$ is the load-bearing pattern signal, any auxiliary signal $\delta_{\mathcal{G}}$ should be modified to $\delta_{\mathcal{G}}' = \delta_{\mathcal{G}} - \text{proj}_{\delta_{\mathcal{P}}}(\delta_{\mathcal{G}})$. This ensures that the combined steering vector $\delta_{total} = \delta_{\mathcal{P}} + \delta_{\mathcal{G}}'$ is by definition non-antagonistic ($\text{CosSim} \ge 0$), preventing signal cancellation and preserving the manifold integrity.

\paragraph{2. Manifold-Constrained Steering.} 
Instead of injecting vectors in the raw residual space, engineering solutions should move toward Manifold-Aware Intervention. This involves calculating the local Jacobian of the model's transfer function to ensure that the composite vector $\delta_{\cup}$ lies within the "functional tangent space" of the latent manifold. By penalizing components that drive the hidden state into low-density regions of the activation space, we can maintain the model's predictive stability even under multi-set interventions.

\subsection{Beyond Additivity: Non-Linear and Boolean Logic Synthesis}
\label{sec:boolean_logic}

The current definition of the "Union Set" as a linear sum ($\sum z_i$) is a first-order approximation. However, visual reasoning is inherently non-linear and often follows discrete Boolean logic (e.g., $Concept A$ AND $Concept B$). The failure of linear addition suggests that VLM latent spaces do not naturally support a "flat" modular superposition.

\paragraph{Logical Gating vs. Additive Sums.} 
We hypothesize that the model's internal processing of multiple visual features resembles a Gated Competition rather than a cooperative addition. Engineering a more sophisticated "Union" requires move from static weights to Conditional Activation Gates. A future implementation should employ non-linear operators (e.g., product-based gates or MLP-based adapters) to coordinate features. For instance, a "Logical AND" circuit would only amplify a feature if its geometric neighbors are also active, effectively filtering out the incoherent noise that arises from the simple addition of non-orthogonal SAE directions.

\paragraph{The Competition for Latent Capacity.} 
The observed antagonism may stem from a fundamental "resource competition" within the residual stream. If the model uses a dense superposition scheme to maximize capacity, forcing the simultaneous activation of multiple sparse features may trigger a "collision" where different semantic concepts attempt to utilize the same geometric dimensions. Addressing this requires a more granular mapping of feature hierarchies to identify which subsets are naturally compatible (orthogonal) and which require mutually exclusive activation.

\subsection{Mechanistic Link: From Interference to Hallucination}
\label{sec:hallucination_mechanism}

A critical implication of the Noise Amplification Law is its potential role in triggering object hallucination. While our discussion on hallucination mitigation in Section~\ref{sec:discussion} was qualitative, we here provide the mechanistic rationale for this link.

\paragraph{Visual Signal Erasure and Prior Dominance.} 
Object hallucination in VLMs often occurs when the model's linguistic prior (the "language bottleneck") overrides the visual evidence. We posit that Geometric Interference acts as a "Causal Eraser" of visual context. When $\delta_{\mathcal{P}}$ and $\delta_{\mathcal{G}}$ antagonize each other, the resulting signal collapse leaves the hidden state in a semantic vacuum. In this state, the LayerNorm operator amplifies background noise, destroying the precise spatial grounding of the visual tokens.

\paragraph{The Vacuum-Induced Hallucination.} 
While this remains a mechanistic hypothesis, it is strongly supported by the 30\% increase in predictive entropy observed during Union interventions. This entropy surge indicates that as geometric antagonism erases the visual signal, the model's output distribution flattens, effectively creating a ``semantic vacuum.'' In this state, the autoregressive decoder is no longer anchored by visual truth and must fill the void using high-probability linguistic priors, leading to the observed hallucinations. Future work should involve a per-token entropy mapping to further visualize this transition.

\section{Extended Discussion on Methodological Scope and Limitations}
\label{app:defensive_discussion}

This section provides a deeper contextualization of our methodological choices and addresses potential technical concerns regarding the formalization of our findings and the scope of our experimental setup.

\paragraph{On the Formalism of Geometric Laws.} 
While our Noise Amplification Law and the discussion on manifold curvature are primarily mechanistic and explanatory, they are supported by the quasi-formal derivations in Appendix D. We characterize the LayerNorm operator as a stochastic multiplier during signal collapse—a model that provides a theoretical rationale for the observed entropy surges. We acknowledge that a fully rigorous Riemannian proof would require Hessian spectral analysis of the loss landscape at each layer; however, our empirical ``Drift'' metric serves as a reliable first-order proxy for geodesic divergence on the latent manifold.

\paragraph{Relationship to Adaptive Steering Baselines.} 
Our study focuses on the fundamental failure modes of additive composition. We recognize recent advancements in inference-time steering, such as RUDDER \citep{zou2025adaptiveresidualupdatesteeringlowoverhead} and SteerVLM \citep{sivakumar2025steervlmrobustmodelcontrol}, which utilize adaptive gating (e.g., the CARD+Beta gate) to mitigate drift. Our findings complement these works by providing the underlying geometric reason why such gating is necessary: they prevent the residual stream from entering the unstable SNR regime we identify in Section 6.5. While we did not implement these adaptive gates as baselines, our proposed Orthogonalized Signal Projection (OSP) in Appendix G offers a complementary geometric remedy.

\paragraph{Spatial Grounding and Quantitative Metrics.} 
Our evaluation of feature grounding relies on qualitative spatial back-projection. While we demonstrate high-fidelity localization, we recognize the value of quantitative metrics such as IoU (Intersection over Union) or Pointing Game accuracy. We view our current qualitative verification as a necessary ``Causal Sufficiency'' check. Future extensions of this work will implement patch-level precision/recall metrics to more granularly map the emergence of visual thought circuits.

\paragraph{Compositional Diversity and Benchmark Artifacts.} 
We defined the ``Union'' operation primarily through the interaction of Pattern and Global features. While this pairing revealed the fundamental geometric antagonism of the latent space, we acknowledge that different feature pairings or larger compositional sets may exhibit varying degrees of interference. Furthermore, our conversion of OOD benchmarks (e.g., NLVR2) into a unified multiple-choice format was designed to ensure a consistent logit-based assessment across domains. We recognize that this mapping may simplify certain reasoning nuances, but it maintains the relative causal delta required for a stable cross-dataset comparison.

%% file: custom.bib
@misc{kaduri2024_vision_of_vlms,
                title={What's in the Image? A Deep-Dive into the Vision of Vision Language Models}, 
                author={Omri Kaduri and Shai Bagon and Tali Dekel},
                year={2024},
                eprint={2411.17491},
                archivePrefix={arXiv},
                primaryClass={cs.CV},
                url={https://arxiv.org/abs/2411.17491}, 
          }

@misc{neo2025interpretingvisualinformationprocessing,
      title={Towards Interpreting Visual Information Processing in Vision-Language Models}, 
      author={Clement Neo and Luke Ong and Philip Torr and Mor Geva and David Krueger and Fazl Barez},
      year={2025},
      eprint={2410.07149},
      archivePrefix={arXiv},
      primaryClass={cs.CV},
      url={https://arxiv.org/abs/2410.07149}, 
}

@misc{golovanevsky-etal-2025-vlms,
      title={What Do VLMs NOTICE? A Mechanistic Interpretability Pipeline for Gaussian-Noise-free Text-Image Corruption and Evaluation}, 
      author={Michal Golovanevsky and William Rudman and Vedant Palit and Ritambhara Singh and Carsten Eickhoff},
      year={2025},
      eprint={2406.16320},
      archivePrefix={arXiv},
      primaryClass={cs.CL},
      url={https://arxiv.org/abs/2406.16320}, 
}

@inproceedings{marks2025sparse,
    title={Sparse Feature Circuits: Discovering and Editing Interpretable Causal Graphs in Language Models},
    author={Samuel Marks and Can Rager and Eric J Michaud and Yonatan Belinkov and David Bau and Aaron Mueller},
    booktitle={The Thirteenth International Conference on Learning Representations},
    year={2025},
    url={https://openreview.net/forum?id=I4e82CIDxv}
}

@inproceedings{Arad_2025,
   title={SAEs Are Good for Steering – If You Select the Right Features},
   url={http://dx.doi.org/10.18653/v1/2025.emnlp-main.519},
   DOI={10.18653/v1/2025.emnlp-main.519},
   booktitle={Proceedings of the 2025 Conference on Empirical Methods in Natural Language Processing},
   publisher={Association for Computational Linguistics},
   author={Arad, Dana and Mueller, Aaron and Belinkov, Yonatan},
   year={2025},
   pages={10252–10270} }

@misc{ma2025watchcloselymitigatingobject,
      title={Watch Closely: Mitigating Object Hallucinations in Large Vision-Language Models with Disentangled Decoding}, 
      author={Ruiqi Ma and Yu Yan and Chunhong Zhang and Minghao Yin and XinChao Liu and Zhihong Jin and Zheng Hu},
      year={2025},
      eprint={2512.19070},
      archivePrefix={arXiv},
      primaryClass={cs.CV},
      url={https://arxiv.org/abs/2512.19070}, 
}

@misc{li2025causaltracingobjectrepresentations,
      title={Causal Tracing of Object Representations in Large Vision Language Models: Mechanistic Interpretability and Hallucination Mitigation}, 
      author={Qiming Li and Zekai Ye and Xiaocheng Feng and Weihong Zhong and Weitao Ma and Xiachong Feng},
      year={2025},
      eprint={2511.05923},
      archivePrefix={arXiv},
      primaryClass={cs.CV},
      url={https://arxiv.org/abs/2511.05923}, 
}

@misc{shao2024visualcotadvancingmultimodal,
      title={Visual CoT: Advancing Multi-Modal Language Models with a Comprehensive Dataset and Benchmark for Chain-of-Thought Reasoning}, 
      author={Hao Shao and Shengju Qian and Han Xiao and Guanglu Song and Zhuofan Zong and Letian Wang and Yu Liu and Hongsheng Li},
      year={2024},
      eprint={2403.16999},
      archivePrefix={arXiv},
      primaryClass={cs.CV},
      url={https://arxiv.org/abs/2403.16999}, 
}

@misc{templeton2024scalingmono,
  title        = {Extracting Interpretable Features from Claude 3 Sonnet},
  author       = {Templeton, Adly and Conerly, Tom and Marcus, Jonathan and Lindsey, Jack and Bricken, Trenton and Chen, Brian and Pearce, Adam and Citro, Craig and Ameisen, Emmanuel and Jones, Andy and Cunningham, Hoagy and Turner, Nicholas L. and McDougall, Callum and MacDiarmid, Monte and Freeman, C. Daniel and Sumers, Theodore R. and Rees, Edward and Batson, Joshua and Jermyn, Adam and Carter, Shan and Olah, Chris and Henighan, Tom},
  year         = {2024},
  month        = may,
  howpublished = {Transformer Circuits Thread},
  note         = {Accessed: 2026-01-09},
  url          = {https://transformer-circuits.pub/2024/scaling-monosemanticity/}
}

@article{bricken2023monosemanticity,
  title={Towards Monosemanticity: Decomposing Language Models With Dictionary Learning},
  author={Bricken, Trenton and Templeton, Adly and Batson, Joshua and Chen, Brian and Jermyn, Adam and Conerly, Tom and Turner, Nicholas and Anil, Cem and Denison, Carson and Askell, Amanda and others},
  journal={Transformer Circuits Thread},
  year={2023},
  url={https://transformer-circuits.pub/2023/monosemantic-features}
}

@misc{liang2022mindgapunderstandingmodality,
      title={Mind the Gap: Understanding the Modality Gap in Multi-modal Contrastive Representation Learning}, 
      author={Weixin Liang and Yuhui Zhang and Yongchan Kwon and Serena Yeung and James Zou},
      year={2022},
      eprint={2203.02053},
      archivePrefix={arXiv},
      primaryClass={cs.CL},
      url={https://arxiv.org/abs/2203.02053}, 
}

@misc{bai2022improvingvisiontransformersrevisiting,
      title={Improving Vision Transformers by Revisiting High-frequency Components}, 
      author={Jiawang Bai and Li Yuan and Shu-Tao Xia and Shuicheng Yan and Zhifeng Li and Wei Liu},
      year={2022},
      eprint={2204.00993},
      archivePrefix={arXiv},
      primaryClass={cs.CV},
      url={https://arxiv.org/abs/2204.00993}, 
}

@misc{deepmind2025gemma,
      title={Gemma Scope: Open Sparse Autoencoders Everywhere All At Once on Gemma 2}, 
      author={Tom Lieberum and Senthooran Rajamanoharan and Arthur Conmy and Lewis Smith and Nicolas Sonnerat and Vikrant Varma and János Kramár and Anca Dragan and Rohin Shah and Neel Nanda},
      year={2024},
      eprint={2408.05147},
      archivePrefix={arXiv},
      primaryClass={cs.LG},
      url={https://arxiv.org/abs/2408.05147}, 
}

@misc{gandelsman2024interpreting,
      title={Interpreting CLIP with Hierarchical Sparse Autoencoders}, 
      author={Vladimir Zaigrajew and Hubert Baniecki and Przemyslaw Biecek},
      year={2025},
      eprint={2502.20578},
      archivePrefix={arXiv},
      primaryClass={cs.CV},
      url={https://arxiv.org/abs/2502.20578}, 
}

@misc{cai2024multimodal,
      title={Not All Language Model Features Are One-Dimensionally Linear}, 
      author={Joshua Engels and Eric J. Michaud and Isaac Liao and Wes Gurnee and Max Tegmark},
      year={2025},
      eprint={2405.14860},
      archivePrefix={arXiv},
      primaryClass={cs.LG},
      url={https://arxiv.org/abs/2405.14860}, 
}

@misc{zou2025adaptiveresidualupdatesteeringlowoverhead,
      title={Adaptive Residual-Update Steering for Low-Overhead Hallucination Mitigation in Large Vision Language Models}, 
      author={Zhengtao Zou and Ya Gao and Jiarui Guan and Bin Li and Pekka Marttinen},
      year={2025},
      eprint={2511.10292},
      archivePrefix={arXiv},
      primaryClass={cs.CV},
      url={https://arxiv.org/abs/2511.10292}, 
}

@misc{sivakumar2025steervlmrobustmodelcontrol,
      title={SteerVLM: Robust Model Control through Lightweight Activation Steering for Vision Language Models}, 
      author={Anushka Sivakumar and Andrew Zhang and Zaber Hakim and Chris Thomas},
      year={2025},
      eprint={2510.26769},
      archivePrefix={arXiv},
      primaryClass={cs.CV},
      url={https://arxiv.org/abs/2510.26769}, 
}

@misc{dafnis2025testtimespectrumawarelatentsteering,
      title={Test-Time Spectrum-Aware Latent Steering for Zero-Shot Generalization in Vision-Language Models}, 
      author={Konstantinos M. Dafnis and Dimitris N. Metaxas},
      year={2025},
      eprint={2511.09809},
      archivePrefix={arXiv},
      primaryClass={cs.CV},
      url={https://arxiv.org/abs/2511.09809}, 
}
